\documentclass[journal]{IEEEtran}
\pdfoutput=1
\usepackage{graphics} 
\usepackage{graphicx}
\usepackage{amsfonts}
\usepackage{amsmath}
\usepackage{amssymb}
\usepackage{url}
\usepackage{multicol}
\usepackage{float}
\usepackage{subfloat}
\usepackage[square, numbers]{natbib}
\usepackage{pgfplots}
\usepackage[font=footnotesize]{caption}
\usepackage{xcolor}
\usepackage[caption=false, font=footnotesize]{subfig}
\usepackage{adjustbox}
\usepackage{array}
\usepackage{booktabs}
\usepackage{makecell}
\usepackage{siunitx}
\usepackage{colortbl}
\usepackage{import}
\usepackage{microtype}
\usepackage{dcolumn}

\renewcommand{\baselinestretch}{0.98}

\newcommand{\etal}{\emph{et~al.}}

\newcommand{\figref}[1]{Fig.~\ref{#1}}
\newcommand{\secref}[1]{Sec.~\ref{#1}}

\renewcommand{\eqref}[1]{Eq.~(\ref{#1})}
\newcommand{\tabref}[1]{Table~\ref{#1}}
\newcommand{\comment}[1]{}
\newcolumntype{P}[1]{>{\centering\arraybackslash}p{#1}}
\newcolumntype{d}[1]{D{.}{.}{#1}}

\usepackage{collcell}
\usepackage{hhline}
\usepackage{pgf}
\usepackage{multirow}

\newcommand\crule[3][black]{\textcolor{#1}{\rule{#2}{#3}}}
\newcommand*\rot{\rotatebox{90}}

\definecolor{asphalt}{RGB}{228, 26, 28}
\definecolor{grass}{RGB}{152, 78, 163}
\definecolor{gravel}{RGB}{55, 126, 184}
\definecolor{parking}{RGB}{77, 175, 74}
\definecolor{cobble}{RGB}{255, 127, 0}

\def\X#1{\raisebox{3.5em}{#1}}

\def\colorModel{hsb} 

\newcommand\ColCell[1]{
  \pgfmathparse{#1<100?1:0}  
    \ifnum\pgfmathresult=0\relax\color{white}\fi
  \pgfmathsetmacro\compA{0}      
  \pgfmathsetmacro\compB{#1/200} 
  \pgfmathsetmacro\compC{1}      
  \edef\x{\noexpand\centering\noexpand\cellcolor[\colorModel]{\compA,\compB,\compC}}\x #1
  } 
\newcolumntype{E}{>{\collectcell\ColCell}m{0.6cm}<{\endcollectcell}}  

\usepgfplotslibrary{colorbrewer}

\begin{document}

\title{Self-Supervised Visual Terrain Classification from Unsupervised Acoustic Feature Learning}
\author{Jannik Z\"urn$^1$, Wolfram Burgard$^{1,2}$, and Abhinav Valada$^1$
\thanks{$^1$Department of Computer Science, University of Freiburg, Germany}\thanks{$^2$Toyota Research Institute, Los Altos, USA.}}

\markboth{\textcopyright 2019 IEEE}%
{ Z\"urn \MakeLowercase{\textit{\etal}}: Self-Supervised Visual Terrain Classification from Unsupervised Acoustic Feature Learning}
\IEEEaftertitletext{\vspace{-1\baselineskip}}

\maketitle

\begin{abstract}
Mobile robots operating in unknown urban environments encounter a wide range of complex terrains to which they must adapt their planned trajectory for safe and efficient navigation. Most existing approaches utilize supervised learning to classify terrains from either an exteroceptive or a proprioceptive sensor modality. However, this requires a tremendous amount of manual labeling effort for each newly encountered terrain as well as for variations of terrains caused by changing environmental conditions. In this work, we propose a novel terrain classification framework leveraging an unsupervised proprioceptive classifier that learns from vehicle-terrain interaction sounds to self-supervise an exteroceptive classifier for pixel-wise semantic segmentation of images. To this end, we first learn a discriminative embedding space for vehicle-terrain interaction sounds from triplets of audio clips formed using visual features of the corresponding terrain patches and cluster the resulting embeddings. We subsequently use these clusters to label the visual terrain patches by projecting the traversed tracks of the robot into the camera images. Finally, we use the sparsely labeled images to train our semantic segmentation network in a weakly supervised manner. We present extensive quantitative and qualitative results that demonstrate that our proprioceptive terrain classifier exceeds the state-of-the-art among unsupervised methods and our self-supervised exteroceptive semantic segmentation model achieves a comparable performance to supervised learning with manually labeled data.
\end{abstract}

\IEEEpeerreviewmaketitle

\section{Introduction}
\label{sec:introduction}

\IEEEPARstart{R}{ecent} advances in robotics and machine learning have enabled the deployment of autonomous robots in challenging outdoor environments for complex tasks such as autonomous driving, last mile delivery, and patrolling. Robots operating in these environments encounter a wide range of terrains from paved roads and cobble stones to unstructured dirt roads and grass. It is essential for them to be able to reliably classify and characterize these terrains for safe and efficient navigation. This is an extremely challenging problem as the visual appearance of outdoor terrain drastically changes over the course of days and seasons, with variations in lighting due to weather, precipitation, artificial light sources, dirt or snow on the ground, among other factors. Therefore, robots should be able to actively perceive the terrains and adapt their navigation strategy as solely relying on pre-existing maps is insufficient.\looseness=-1

These factors have motivated substantial research in learning to classify terrains, both using exteroceptive~\cite{sofman2006improving, hadsell2008deep, konolige2009mapping} or proprioceptive~\cite{brooks2012self, ojeda2006terrain, valada2018deep} sensor modalities. Proprioceptive sensors sense terrain properties through the interaction of the robot with its environment and their data can be used to train accurate terrain classifiers~\cite{brooks2012self}. Among the various proprioceptive modalities, vehicle-terrain interaction sounds from mobile robots in particular have been shown to have highly distinctive features that strongly correlate with the underlying semantic terrain classes and thereby enable fine-grained terrain classification~\cite{ojeda2006terrain, libby2012using, valada2017deep}. Exteroceptive sensors, in contrast, sense the terrain from a distance and enable a robot to classify its surroundings without directly interacting with it. Learning from the combination of proprioceptive and exteroceptive sensor modalities allows us to associate terrain features in the vicinity of the robot to more distant features that are ahead of the robot. We also reason that it is extremely difficult to capture properties of all the different terrains as well as their variations due to changing environmental conditions, using just one modality. Recent works in multimodal deep learning~\cite{xie2019best, radwan2018multimodal, blum2018modular, valada2016towards} have demonstrated the ability to learn robust complementary features which yield superior performance in many learning tasks. We follow this paradigm and leverage two diverse modalities, sound and vision, to learn complementary features for robust terrain classification.\looseness=-1

\begin{figure}
\centering
\includegraphics[width=0.99\linewidth, trim={10cm 5cm 20cm 0cm}, clip]{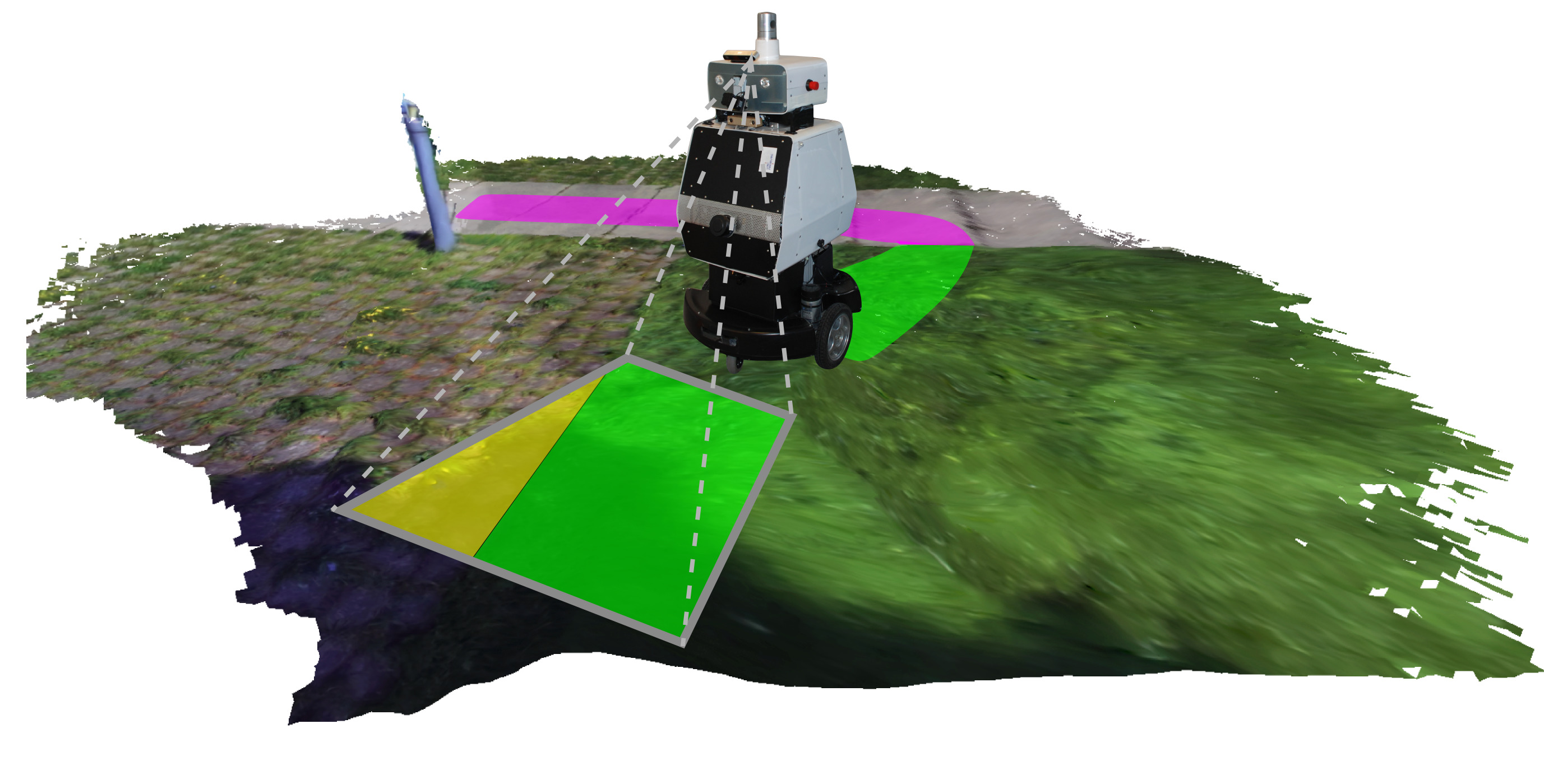}
\caption{Our self-supervised approach enables a robot to classify urban terrains without any manual labeling using an on-board camera and a microphone. Our proposed unsupervised audio classifier automatically labels visual terrain patches by projecting the traversed tracks into camera images. The resulting sparsely labeled images are used to train a semantic segmentation network for visually classifying new camera images in a pixel-wise manner.}
\label{fig:coverimage}
\end{figure}

Most state-of-the-art learning methods require a significant amount of data samples which are often arduous to obtain in supervised learning settings where labels have to be manually assigned to data samples. Moreover, these models tend to degrade in performance once presented with data sampled from a distribution that is not present in the training data. In order to perform well on data from a new distribution, they have to be retrained after repeated manual labeling which in general is unsustainable for widespread deployment of robots. Self-supervised learning allows the training data to be labeled automatically by exploiting the correlations between different input signals thereby reducing the amount of manual labelling work by a large margin. Furthermore, unsupervised audio classification eliminates the need to manually label audio samples. We take a step towards lifelong learning for visual terrain classification by leveraging the fact that the distribution of terrain sounds does not depend on the visual appearance of the terrain. This enables us to employ our trained audio terrain classification model in previously unseen visual perceptual conditions to automatically label patches of terrain in images, in a completely self-supervised manner. The visual classification model can then be fine-tuned on the new training samples by leveraging transfer learning to adapt to the new appearance conditions.\looseness=-1

In this work, we present a novel self-supervised approach to visual terrain classification by exploiting the supervisory signal from an unsupervised proprioceptive terrain classifier utilizing vehicle-terrain interaction sounds. \figref{fig:coverimage} illustrates our approach where our robot equipped with a camera and a microphone traverses different terrains and captures both sensor streams along its trajectory. The poses of the robot recorded along the trajectory enables us to associate the visual features of a patch of ground that is in front of the robot initially with its corresponding auditory features when that patch of ground is traversed by the robot. We split the audio stream into small snippets and embed them into an embedding space using metric learning. To this end, we propose a novel triplet sampling method based on the visual features of the respective terrain patches. This now enables the usage of triplet loss formulations for metric learning without requiring ground truth labels. We obtained the aforementioned visual features from an off-the-shelf image classification model pre-trained on the ImageNet dataset. To the best of our knowledge, our work is the first to exploit embeddings from one modality to form triplets for learning an embedding space for samples from an extremely different modality. We interpret the resulting clusters formed by the audio embeddings as labels for training a weakly-supervised visual semantic terrain segmentation model. We then employ this model for pixel-wise classification of terrain that is in front of the robot and use this information for trajectory planning.

In order to facilitate this work, we collected a large-scale urban terrains dataset consisting of five terrain categories in diverse environments at different times of the day and varying weather conditions. We present extensive quantitative and qualitative evaluations of our framework that demonstrate that our unsupervised proprioceptive terrain classifier achieves state-of-the-art performance for unsupervised terrain classification from vehicle-terrain interaction sounds and our self-supervised visual terrain semantic segmentation model achieves a comparable performance to supervised learning. More importantly, we also show that exploiting the training signal from proprioceptive terrain classification for self-supervised exteroceptive semantic segmentation enables our robot to learn a robust terrain classification model that can adapt to changes in visual perception caused by external sources such as illumination, time-of-day, weather conditions or seasonal changes. Thereby taking a step towards lifelong learning of traversability estimation.

In summary, the major contributions of this work are:
\begin{itemize}
\item A general terrain classification framework for mobile robots that uses an unsupervised proprioceptive classifier to self-supervise an exteroceptive classifier.
\item A novel heuristic to form triplets for metric learning that does not require ground truth labels but instead leverages a complementary modality.
\item A self-supervised visual semantic segmentation model that learns from weakly labeled bird's eye view images.
\item The new \textit{Freiburg Terrains} dataset consisting of more than four hours of audio-video recordings of terrain traversals tagged with SLAM poses.
\item Extensive quantitative as well as qualitative evaluations and ablation studies demonstrating the effectiveness of our proposed framework.
\end{itemize}

The remainder of this paper is organized as follows. In~\secref{sec:related}, we discuss related work on self-supervised and semi-supervised terrain classification using proprioceptive and exteroceptive sensors. We then describe our pre-processing pipeline, followed by our unsupervised acoustic feature learning approach and our self-supervised visual terrain classification method in \secref{sec:approach}. In \secref{sec:experiments}, we present extensive empirical evaluations with ablation studies followed by a discussion in \secref{sec:conclusions}.
\section{Related Work}
\label{sec:related}

Self-supervised learning of terrain classification and terrain properties for mobile robots has been investigated intensively in recent years. Early works by Sofman~\etal~\cite{sofman2006improving} propose a self-supervised online learning approach that relies on overhead imagery such as satellite images to learn a traversability costmap for outdoor off-road robots. Happold~\etal~\citep{happold2006enhancing} train a neural network offline on hand-labeled geometric features computed from stereo data for online traversability analysis using the predictions of the trained network. Later works by Hadsell~\etal~\citep{hadsell2008deep, hadsell2009learning} and Konolige~\etal~\citep{konolige2009mapping} demonstrated early success in long-range terrain classification using a deep belief network on the LAGR robot platform. Hadsell~\etal~\citep{hadsell2009learning} present an obstacle detection and path detection approach for outdoor environments based exclusively on stereo vision and self-supervised learning. They trained a deep belief network on terrain imagery using labels obtained from a short-range stereo vision classifier for differentiating between multiple traversability classes such as \textit{Ground}, \textit{Obstacle} or \textit{Footline}. Using the same robot platform, Konolige~\etal~\cite{konolige2009mapping} create pixel-wise texture statistics in a small neighborhood around each pixel. During classification, each pixel is assigned to a cluster of histograms with similar properties using the euclidean distance in histogram space. However, these approaches have only been demonstrated on a small number of terrain classes and they do not exploit complementary proprioceptive modalities to improve accuracy.\looseness=-1

The use of multimodal sensor data for self-supervised learning has also been investigated in several works~\cite{otsu2016autonomous, bekhti2014terrain, brooks2012self, zhou2012self, stavens2012self, otsu2016autonomous, wellhausen2019should, nava2019learning}. Typically, proprioceptive modalities such as vibrations are combined with exteroceptive modalities such as visual images. Brooks~\etal~\cite{brooks2012self} use a proprioceptive vibration sensor to classify the type of terrain their wheeled robot traverses and an exteroceptive vision-based sensor to classify terrain in the field of view in front of the robot. Bekhti~\etal~\citep{bekhti2014terrain} introduced a learning-free scheme to find the correlation between exteroceptive image observations and proprioceptive acceleration signals for the assessment of terrain maneuverability for mobile robots. They perform Canonical Correlation Analysis to compare two groups of quantitative variables to determine if they describe the same type of terrain. Existing work in this area is primarily focused on exploiting vibration-based sensors for proprioceptive terrain classification. Vibration characteristics from terrains are often extremely susceptible to shaking and vibrations from the robot platform itself which often leads to misclassifications.

Previous works in metric learning~\citep{chopra2005learning, schroff2015facenet, sohn2016improved} are targeted towards learning easily clusterable embeddings using loss functions that enable the neural encoders to embed inputs with the same class close to each other and inputs with different classes far from each other. The most common attribute among them is that they propose a loss function that encourages the network to learn favorable embedding spaces that highly correlate with the ground truth labels. Therefore, they rely on ground truth labeling for forming contrastive data tuples that allow the network to learn such embeddings. Other approaches~\citep{xie2016unsupervised} and the subsequent work by Gou~\etal~\citep{guo2017improved} focus on improving the learned embedding spaces from an autoencoder by defining a centroid-based probability distribution and minimizing its Kullback-Leibler divergence to an auxiliary target distribution. This simultaneously improves clustering assignment and the feature representations. They do not rely on any ground truth data and their approach can be leveraged in a plug-in fashion for improving given embeddings without supervision. While these methods are capable of generating embeddings that are more accurately clusterable than plain autoencoders, they do not explore how complementary data from a different modality can be incorporated into the model training to improve the performance. Moreover, these methods only improve the clustering accuracies of embeddings by a small margin compared to plain autoencoders.

In recent years, deep learning based approaches to semantically segment scenes have increasingly been leveraged for visual terrain classification. Barnes~\etal~\citep{barnes2017find} use the trajectory of a manually driven vehicle for self-supervised labeling of urban scenes to detect drivable areas. The vehicle trajectory is used for implicit labeling of pixels in a scene. Due to the lack of additional semantic information, their approach is limited to differentiating between a proposed path, drivable areas, and obstacles. Hirose~\etal~\citep{hirose2018gonet} recently presented a semi-supervised deep learning approach to traversability estimation from fisheye camera images. They leverage Generative Adversarial Networks to effectively predict whether the area seen in the input images is safe for a robot to traverse. Nava~\etal~\citep{nava2019learning} introduce a self-supervised approach to predict future outputs of a short-range proximity sensor based on the current outputs of a long-range sensor, thus, training a Convolutional Neural Network (CNN) to accurately predict obstacles in the camera images. More recently, Wellhausen~\etal~\cite{wellhausen2019should} proposed a self-supervised terrain property learning method that uses the proprioceptive force-torque signal of a quadrupedal robot as a sparse label generator for the supervised training of a CNN for semantic segmentation of camera images. The force-torque signal of the robots legs at the foothold positions are classified using a ground reactive score and serve as sparse ground truth signals for supervised semantic segmentation. While recent works have demonstrated progress towards self-supervised terrain-classification, they either do not effectively generalize to different classes~\cite{barnes2017find} or do not leverage the complementary nature of multimodal perception for different terrain classes~\cite{wellhausen2019should}.

Inspired by these recent works, we demonstrate the benefits of self-supervised multimodal terrain classification by fully exploiting the complementary nature of proprioception and exteroception. Specifically, our framework does not require any manual labeling of data samples and we show that automatic data collection with a robot coupled with the self-labeling steadily improves the performance as well the robustness of our model. More importantly, our work takes a step towards lifelong learning of traversability estimation by reusing our unsupervised audio classifier to adapt our visual terrain classifier in adversary lighting conditions that are not present in the training data.
\section{Technical Approach}
\label{sec:approach}

\begin{figure*}
\centering
\includegraphics[width=\linewidth]{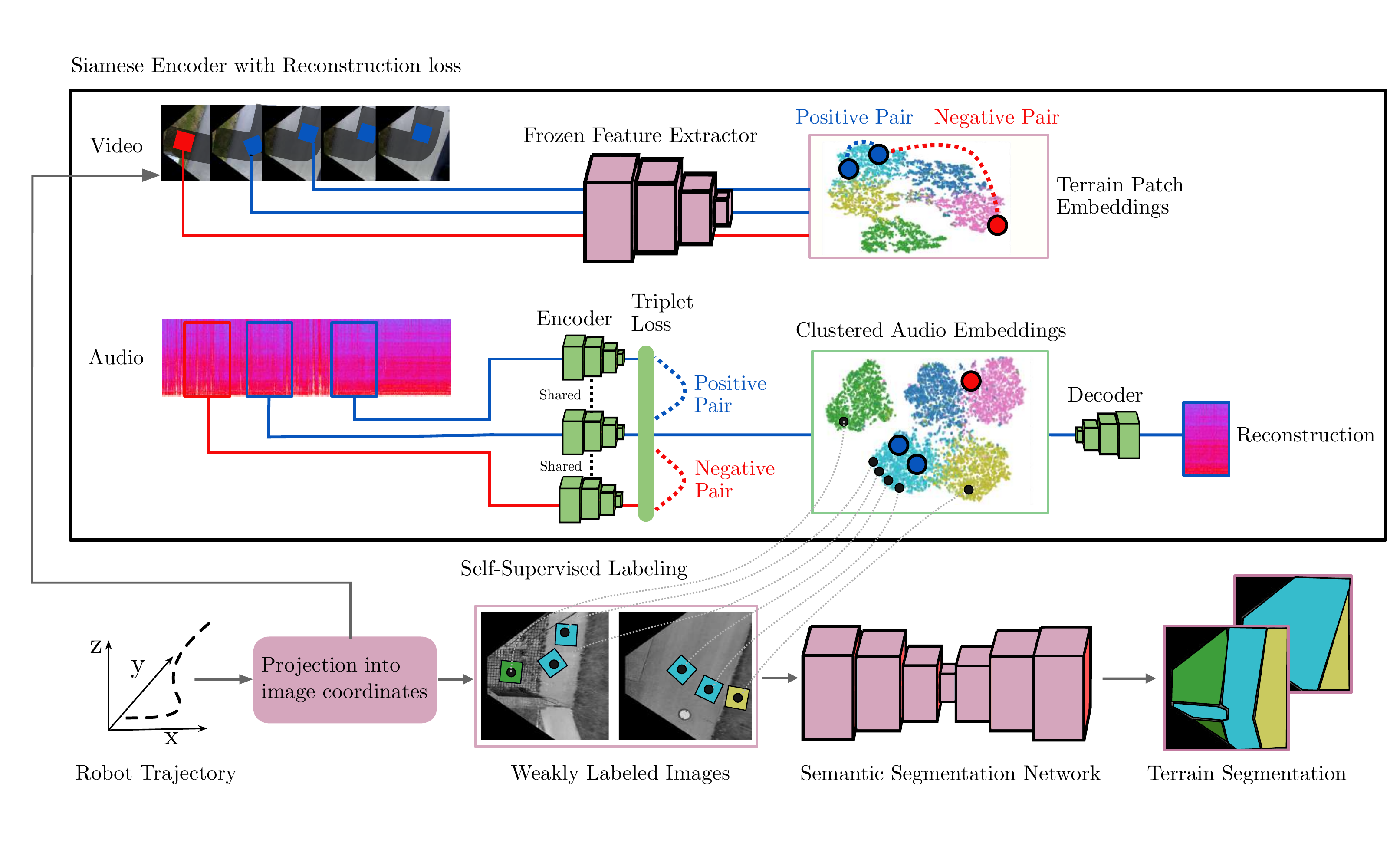}
\caption{Overview of our proposed self-supervised terrain classification framework. The upper part of the figure illustrates our novel Siamese Encoder with Reconstruction loss (SE-R), while the lower part illustrates how the labels obtained from the SE-R are used to automatically annotate data for training the semantic segmentation network. The camera images are first projected into a birds-eye-view perspective of the scene and the trajectory of the robot is projected into this viewpoint. In our SE-R approach, using both the audio clips from the recorded terrain traversal and the corresponding patches of terrain recorded with a camera, we embed each clip of the audio stream into an embedding space that is highly discriminative in terms of the underlying terrain class. This is performed by forming triplets of audio samples using the visual similarity of the corresponding patches of ground obtained with a pre-trained feature extractor. We then cluster the resulting audio embeddings and use the cluster indices as labels for self-supervised labeling. The resulting labeled images serve as a weakly labeled training dataset for a semantic segmentation network for pixel-wise terrain classification.}
\label{fig:approach}
\end{figure*}

In this section, we detail our proposed self-supervised terrain classification framework. \figref{fig:approach} visualizes the overall information flow in our system. While acquiring the images and audio data, we tag each sample with the robot pose obtained using our SLAM system~\cite{kummerle2015autonomous}. We then project the camera images into a birds-eye-view perspective and project the path traversed by the robot in terms of its footprint into this viewpoint. We transform the audio clips into a spectrogram representation and embed them into an embedding space using our proposaed Siamese Encoder with Reconstruction loss (SE-R) on audio triplets that uses features in the visual domain for triplet forming. Subsequently, we cluster the embeddings and use the cluster indices to automatically label the corresponding robot path segments in the birds-eye-view images. The resulting labeled images serve as weakly labeled training data for the semantic segmentation network. Note that the entire approach is executed completely in an unsupervised manner. The cluster indices can be used to indicate terrain labels such as \textit{Asphalt} and \textit{Grass} or in terms of terrain properties.

In the rest of this section, we detail each of the aforementioned components of our framework. In \secref{sec:preprocessing}, we first describe the audio pre-processing that we employ to convert the raw audio signal into audio spectrograms. We then present the approach to project the recorded robot path into the camera frames in \secref{sec:projection}. In the subsequent \secref{sec:featurelearning}, we detail our novel Siamese Encoder variants for unsupervised clustering of audio samples. Finally, in \secref{sec:terrainclassification}, we describe our approach to self-supervised semantic segmentation of camera images using the weak labels obtained from the unsupervised audio clustering.

\subsection{Audio Preprocessing}
\label{sec:preprocessing}

We first split the audio stream into short clips of time window length $t_w$. We then convert each clip into its spectrogram representation using the Short Time Fourier Transform (STFT). Denoting $x$ as the raw audio signal of a clip, the discrete-time STFT generates a two-dimensional representation of the signal as\looseness=-1
\begin{equation}
\text{STFT}\{x[n]\}(m, \omega) = \sum_{n=-\infty}^{\infty} x[n] w[n-m] e^{-j \omega n},
\end{equation}
where $x[n]$ denotes the time-discrete signal and $w[n]$ denotes the window size for the Fast-Fourier-Transform. Finally, the squared magnitude of the complex-valued STFT yields the spectrogram representation of the audio signal.

\subsection{Trajectory Projection Into Image Coordinates}
\label{sec:projection}

We record the stream of monocular camera images from an on-board camera and the corresponding audio stream of the vehicle-terrain interaction sounds from a microphone mounted near the wheel of our robot. We project the robot trajectory into the image coordinates using the robot poses obtained using our SLAM system~\cite{kummerle2015autonomous}. In contrast to other works such as \cite{wellhausen2019should}, we additionally perform perspective warping of the camera images in order to obtain a birds-eye view representation. The robot trajectory is interpreted as a curve $\mathbf{x} \in \mathbb{R}^3$. Denoting the intrinsic camera calibration matrix as $\mathbf{K}$, the perspective transformation matrix as $\mathbf{P}$, the transformation matrix from global coordinates into the camera coordinate frame at time $t$ as $\mathbf{T}^t$, the robot trajectory $\mathbf{u}^t$ at time $t$ in image coordinates can be expressed as
\begin{equation}
\mathbf{u}^t = \mathbf{P} \mathbf{K}  \mathbf{T}^t \mathbf{x}.
\end{equation}

The footprint of the robot entailing its four wheels is circular with a radius of $\SI{0,4}{\meter}$. Every area of the ground covered by the footprint of the robot is considered as being traversed by the robot. Thus, we extend the area that is considered as traversed by $\SI{0,4}{\meter}$ on both sides of the robot trajectory. We denote this area as the \textit{robot path}. We label the areas of ground overlapping with the robot path in every image using our self-supervised approach. The remaining pixels are unlabeled as the robot never traversed the corresponding patches of ground and thus no information about the terrain in such areas is available.\looseness=-1

\subsection{Unsupervised Acoustic Feature Learning}
\label{sec:featurelearning}

Each terrain patch that the robot traverses is represented by two modalities: audio and vision. We obtain the visual representation of a terrain patch from a distance using an on-board camera, while we record the vehicle-terrain interaction sounds by traversing the corresponding terrain patch. For our unsupervised acoustic feature learning approach, we exploit the highly discriminative visual embeddings of terrain patch images obtained using a CNN pre-trained on the ImageNet dataset to form triplets of audio samples. To form such discriminative clusters of embeddings, triplet losses have been proposed~\cite{schroff2015facenet}. We argue that the relative position of terrain patch image embeddings in embedding space serve as a good approximation for ground truth labels that have previously been relied on for triplet forming. We form triplets of audio clips using this heuristic. Finally, we train a Siamese Encoder (SE), and a variant with an additional reconstruction loss (SE-R) in order to embed these audio clips into a highly discriminative audio embedding space.

A triplet $T$ consists of anchor sample $x$, a positive sample $x^+$ with the same class as $x$ and a negative sample $x^-$ with a class different from $x$. Denoting $\mathcal{T}$ as the as the set of triplets, the triplet loss objective can be formulated as

\begin{equation}
\mathcal{L}_{t} = \frac{1}{|\mathcal{T}|} \sum_{(i,j,k) \in \mathcal{T}}\max[D_{i,j}^2 + \alpha - D_{i,k}^2 , 0]
\end{equation}
where $D_{i,j} = ||x_i - x_j||^2$ denotes the euclidean distance between two embeddings $x_i$ and $x_j$. The parameter $\alpha$ denotes a margin between the positive samples and the negative samples. In all our experiments, we use $\alpha = 1$. The triplet loss enforces that the embeddings of samples with the same label are pulled together in embedding space and embeddings of samples with different labels are pushed away from each other simultaneously. As the ground truth labels of the audio samples are not available to form triplets, we argue that an unsupervised heuristic can serve as a substitute signal for the ground truth labels for triplet creation: the local neighborhood of the terrain image patch embeddings. We obtain rectangular patches of terrain by selecting segments of pixels along the robot path, as illustrated in \figref{fig:weak-labels}.\looseness=-1

The closest neighbor in the embedding space has a high likelihood of belonging to the same ground truth class as the anchor sample. Therefore, for sampling triplets, we select the sample with the smallest euclidean distance in the visual embedding space as a positive sample. We then select negative samples by randomly selecting samples that are in a different cluster in visual embedding space than the anchor sample. Although it cannot be always guaranteed that the negative sample does not have the same ground truth class, it has a high likelihood of belonging to a different class, which we observe in our experiments. Likewise, we argue that visually similar terrain patches have high likelihood of belonging to the same class. This means that in practice a fraction of the generated triplets are not correctly formed. However, we empirically find that it is sufficient if the majority all triplets have correct class assignments as they outweigh the incorrectly defined triplets.

We train a Siamese Encoder on the spectrogram representations of the terrain audio clips, as described in \secref{sec:preprocessing}. We use the architecture proposed in~\citep{engel2017neural} as a strong baseline for embedding audio samples. The objective of the Siamese Encoder is to minimize the triplet loss $\mathcal{L}_{t}$. In addition to the Siamese Encoder variant, we propose to add a convolutional decoder that transforms the embeddings back into the original spectrogram domain and impose a reconstruction loss on the decoded samples. We denote this variant as SE-R. We formulate the reconstruction loss $\mathcal{L}_{r}$ as
\begin{equation}
\mathcal{L}_{r} = ||g(f(x_i)) - x_i||^2,
\end{equation}
where  $f$ and $g$ denote the encoder and decoder, respectively.

The objective of the SE-R model is to simultaneously minimize a weighted sum of the triplet loss and the reconstruction loss. Thus the complete loss objective can be formulated as 
\begin{equation}
\mathcal{L} = \beta \mathcal{L}_{t} + (1-\beta) \mathcal{L}_{r}
\end{equation}
where $\beta$ denotes a weighting factor between the two loss components. Setting $\beta = 0$ results in a pure reconstruction loss, while setting $\beta = 1$ results in only optimizing the triplet loss, which is equivalent to our SE approach. For our SE-R approach, we chose $\beta = 0.5$. Finally, we perform k-means clustering of the embeddings to separate the samples into $K$ clusters, corresponding to the $K$ terrain classes present in the dataset. Our approach only requires us to set the number of terrain classes that are present and assign terrain class names to the cluster indices.

\subsection{Self-Supervised Visual Terrain Classification}
\label{sec:terrainclassification}

\begin{figure}
\centering
\includegraphics[width=0.85\linewidth, trim={1cm 0cm 0cm 0cm},clip]{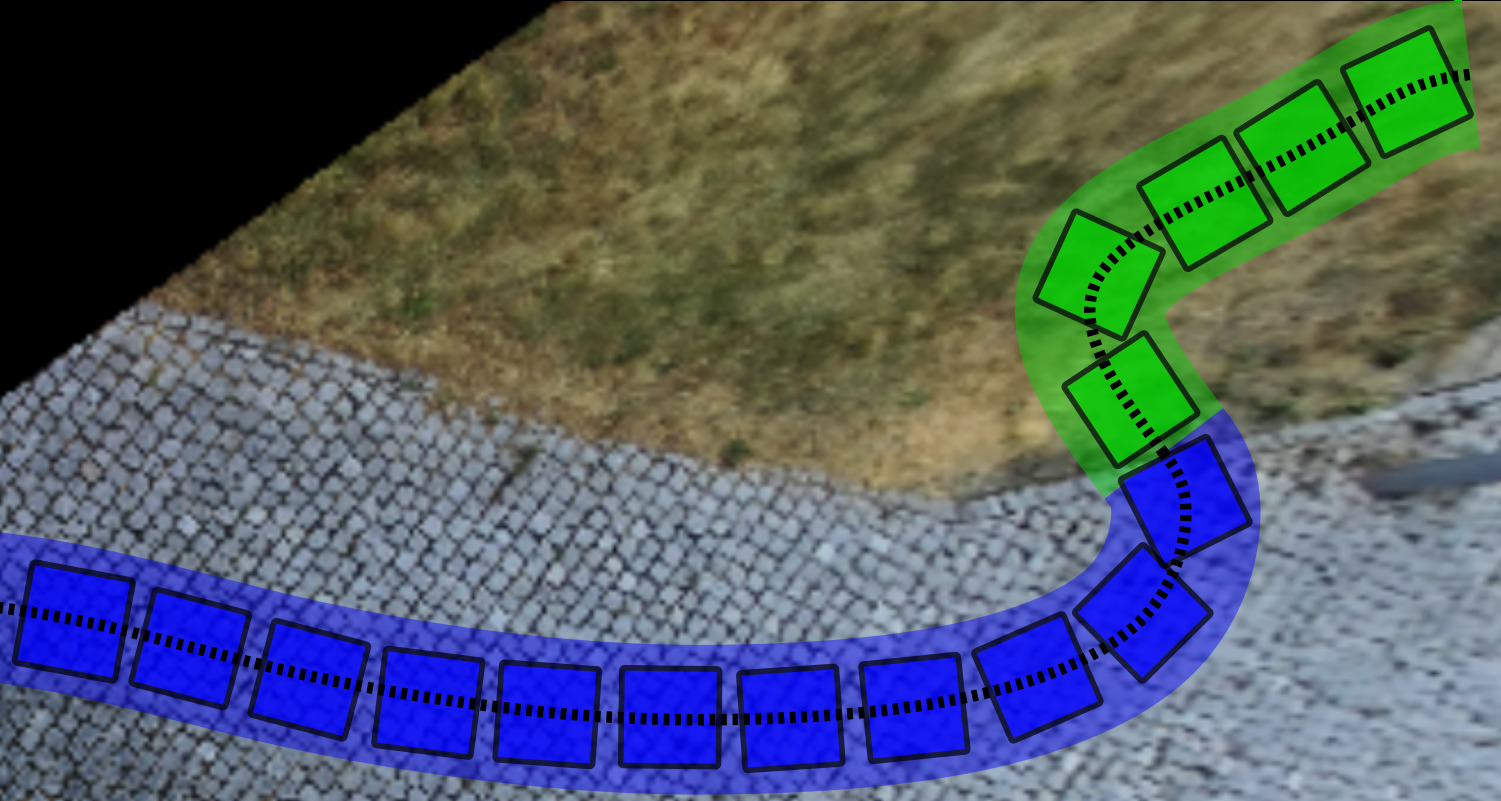}
\caption{An example birds-eye-view of the terrain with the robot path superimposed. Pixels with known class \textit{Cobblestone} and \textit{Grass} are indicated in blue and green respectively. The colored rectangles represent patches of terrain whose features are used to form audio sample triplets. No labels are available for the remaining pixels as the robot has not traversed those regions.}
\label{fig:weak-labels}
\end{figure}

We use the resulting weakly self-labeled birds-eye-view scenes to train a semantic segmentation network in a self-supervised manner. A self-supervisory signal can be obtained for every image pixel that contains a part of the robot path for which the label is known from the unsupervised audio clustering. Note that the segmentation masks for the traversed terrain types are incomplete as the robot cannot be expected to traverse every physical location of terrain in the view to generate complete segmentation masks. We alleviate this challenge by considering all the pixels in camera images that do not contain the robot path as a background class that does not contribute to the segmentation loss. \figref{fig:weak-labels} illustrates a typical training image with its segmentation mask superimposed. We deal with the class imbalance in the training set by weighing each class proportional to its log frequency in the training data set.\looseness=-1

\begin{figure*} 
\captionsetup[subfigure]{labelformat=empty}
    \centering
  \subfloat[]{%
       \includegraphics[width=0.19\linewidth]{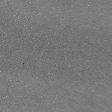}}
    \hfill
   \subfloat[]{%
       \includegraphics[width=0.19\linewidth]{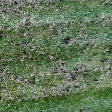}}
    \hfill
  \subfloat[]{%
        \includegraphics[width=0.19\linewidth]{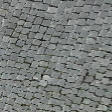}}
    \hfill
  \subfloat[]{%
        \includegraphics[width=0.19\linewidth]{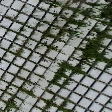}}
    \hfill
    \subfloat[]{%
        \includegraphics[width=0.19\linewidth]{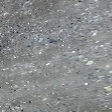}}
\vspace{-0.6cm}
          \subfloat[(a) Asphalt\label{1a}]{%
       \includegraphics[width=0.19\linewidth, trim={3cm 3cm 3cm 4cm},clip]{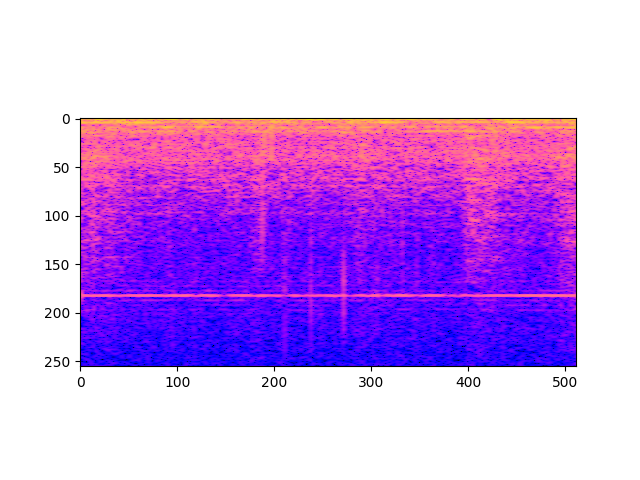}}
    \hfill
   \subfloat[(b) Grass\label{1b}]{%
       \includegraphics[width=0.19\linewidth, trim={3cm 3cm 3cm 4cm},clip]{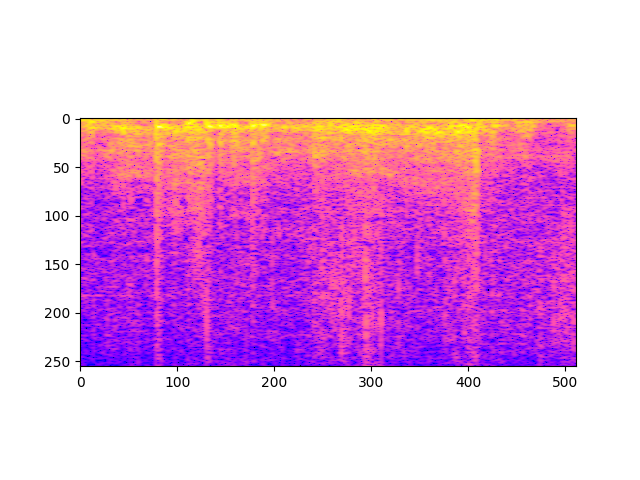}}
    \hfill
  \subfloat[(c) Cobblestone\label{1c}]{%
        \includegraphics[width=0.19\linewidth, trim={3cm 3cm 3cm 4cm},clip]{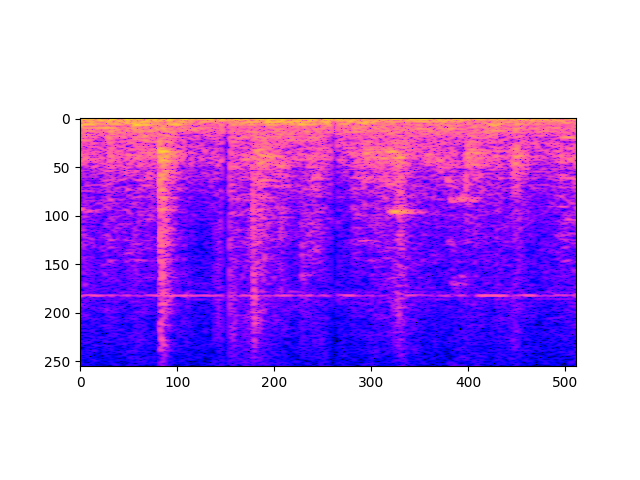}}
    \hfill
  \subfloat[(d) Parking Lot\label{1d}]{%
        \includegraphics[width=0.19\linewidth, trim={3cm 3cm 3cm 4cm},clip]{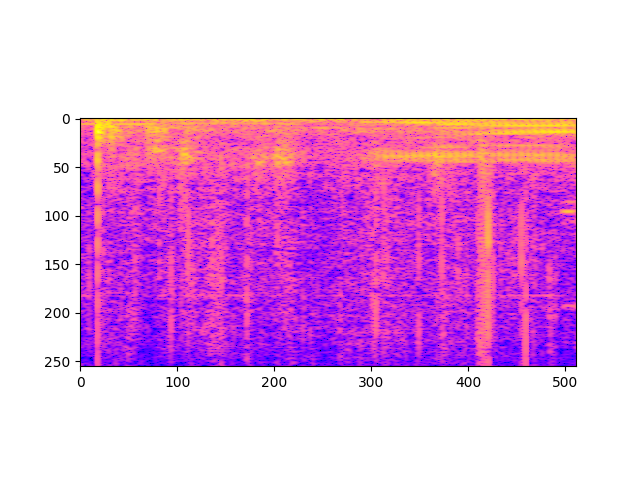}}
    \hfill
    \subfloat[(e) Gravel\label{1e}]{%
        \includegraphics[width=0.19\linewidth, trim={3cm 3cm 3cm 4cm},clip]{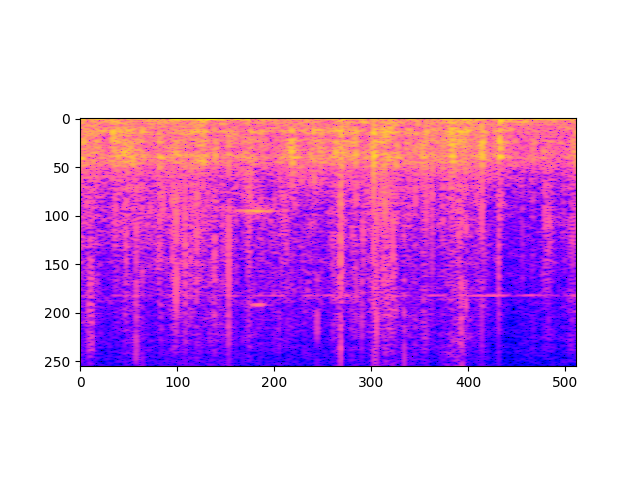}}
  \caption{Example birds-eye-view images and the corresponding spectrogram of the vehicle-terrain interaction sounds from the five different terrain classes present in our dataset. Some terrains have visually similar colors and texture (\textit{a} and \textit{e}), while some terrains also exhibit similar audio signatures (\textit{b} and \textit{e}). Several sequences also contain illumination changes, as well as different terrains bleeding into one another (in \textit{d} where grass grows between stone tiles).}
  \label{fig:classes} 
\end{figure*}

The segmentation loss formulation for one image using class weighted cross-entropy is defined as
\begin{equation}\label{eq:loss}
\mathcal{L}_{seg} = - \sum_{i} \sum_{c}^C w_c y_{i,c} \log ( p_{i,c} )
\end{equation}
where $w_c$ is the respective weight for each class, $y_{i,c}$ denotes the ground truth label at image coordinate $i$, and $\log ( p_{i,c} )$ denotes the log probability from the $\mathsf{softmax}$ output layer of the neural network. 
\section{Experimental Evaluation}
\label{sec:experiments}

We evaluated our framework exhaustively on real-world data that was collected using our Obelix robot platform in diverse environments and perceptual conditions. In this section, we first describe the methodology that we used for collecting our Freiburg Terrains dataset in \secref{sec:freiburgTerrains}, followed by a description of the standard evaluation metrics that we use to quantify the performance in \secref{sec:evaluation-metrics} and the training procedure that we employ in \secref{sec:network-training}. We then present experimental results of our unsupervised acoustic feature clustering approach in \secref{sec:clustering-results} and evaluations of our self-supervised semantic segmentation in \secref{sec:sss}. In \secref{sec:qualitative}, we present qualitative analysis in challenging scenarios and detailed ablation studies in \secref{sec:ablation}. Finally we present results that demonstrate the generalization ability of our model in \secref{sec:generalization} and its utility for trajectory planning in \secref{sec:planning}.

\subsection{Dataset}
\label{sec:freiburgTerrains}

We collected a large-scale dataset using our Obelix robot equipped with a ZED stereo camera mounted on top and pointing downwards onto the ground at an angle of $\ang{30}$ from the horizon. Images were captured at a frequency of approximately $\SI{2}{\Hz}$. For capturing vehicle-terrain interaction sounds, we equipped our robot with a Rode Video-Mic directional microphone that we mounted close to the rear wheel of the robot, pointing towards the contact area between wheel and ground terrain. We captured the audio data at a sampling rate of $\SI{44100}{\Hz}$ and a bit-depth of $\SI{16}{\bit}$. We then split the audio stream into small clips of $\SI{500}{\ms}$ and tagged each clip as well as the time-synchronized images with the pose of the robot that we obtained using our SLAM system~\cite{kummerle2015autonomous}. During the data collection runs, we also varied the speed of the robot from $\SI[per-mode=symbol]{0,2}{\meter\per\second}$ to $\SI[per-mode=symbol]{1,0}{\meter\per\second}$ to capture diverse variations in the vehicle-terrain interaction sounds.

As our robot is equipped with rubber wheels, it is capable of traversing smooth to rough hard surfaces such as asphalt, cement, cobblestone or sett paving. It is also able to slowly traverse off-road terrains such as mowed-grass or gravel paths with hard surfaces. However, it is not suitable for traversing wet mud, crushed stone or puddles due to the large risk of toppling over or leading to entrenchment of the wheels. Therefore, we chose to collect data on five different terrains, namely, \textit{Asphalt}, \textit{Grass}, \textit{Cobblestone}, \textit{Parking Lot}, and \textit{Gravel}. \figref{fig:classes} shows example images of these terrains along with their corresponding spectrograms of the vehicle-terrain interaction sounds. As we see, some of these classes have very similar visual appearance such as \textit{Asphalt} and \textit{Gravel}, while some of the other classes such as \textit{Grass} and \textit{Parking Lot} have very similar auditory features.\looseness=-1

Our dataset contains over four hours of audio and video recordings that were collected over the course of three weeks. In order to have diverse conditions in the dataset, we carried out the data collection runs at different times of day as well as in varying weather conditions including sunny, cloudy, and overcast. Our dataset also contains diverse audio conditions due to wind and distant ambient noises such as construction work. To quantify the performance of our models, we manually labeled the robot path segments with the respective ground truth terrain class. This yields a set of weakly labeled images as only the pixels of the image that contain the path segments navigated by the robot are labeled. Furthermore, we manually densely annotated 200 birds-eye-view terrain images in a pixel-wise manner to yield the \textit{fully labeled} validation set that we also use in our experiments. Additionally, we evaluate our approach on a complementary \textit{weakly labeled} evaluation set containing 500 images that are neither present in the training set or the fully labeled validation set.

\subsection{Evaluation Metrics}
\label{sec:evaluation-metrics}

\begin{figure*} 
\captionsetup[subfigure]{labelformat=empty}
    \centering
  \subfloat[(a) Epoch 0\label{e0}]{%
       \includegraphics[width=0.25\linewidth, trim={6.5cm 4cm 5cm 4cm},clip]{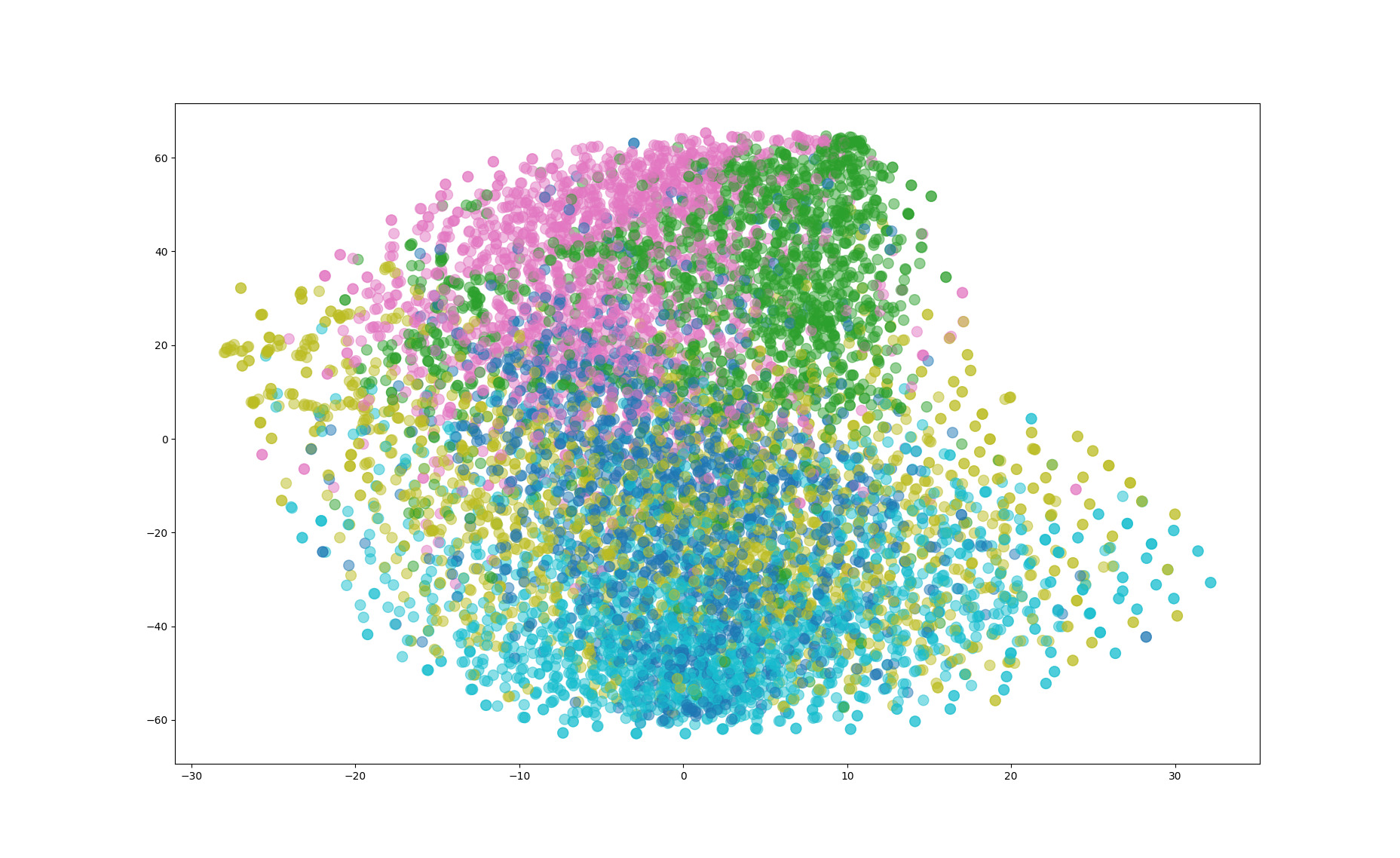}}
    \hfill
   \subfloat[(b) Epoch 10\label{e20}]{%
       \includegraphics[width=0.25\linewidth, trim={6.5cm 4cm 5cm 4cm},clip]{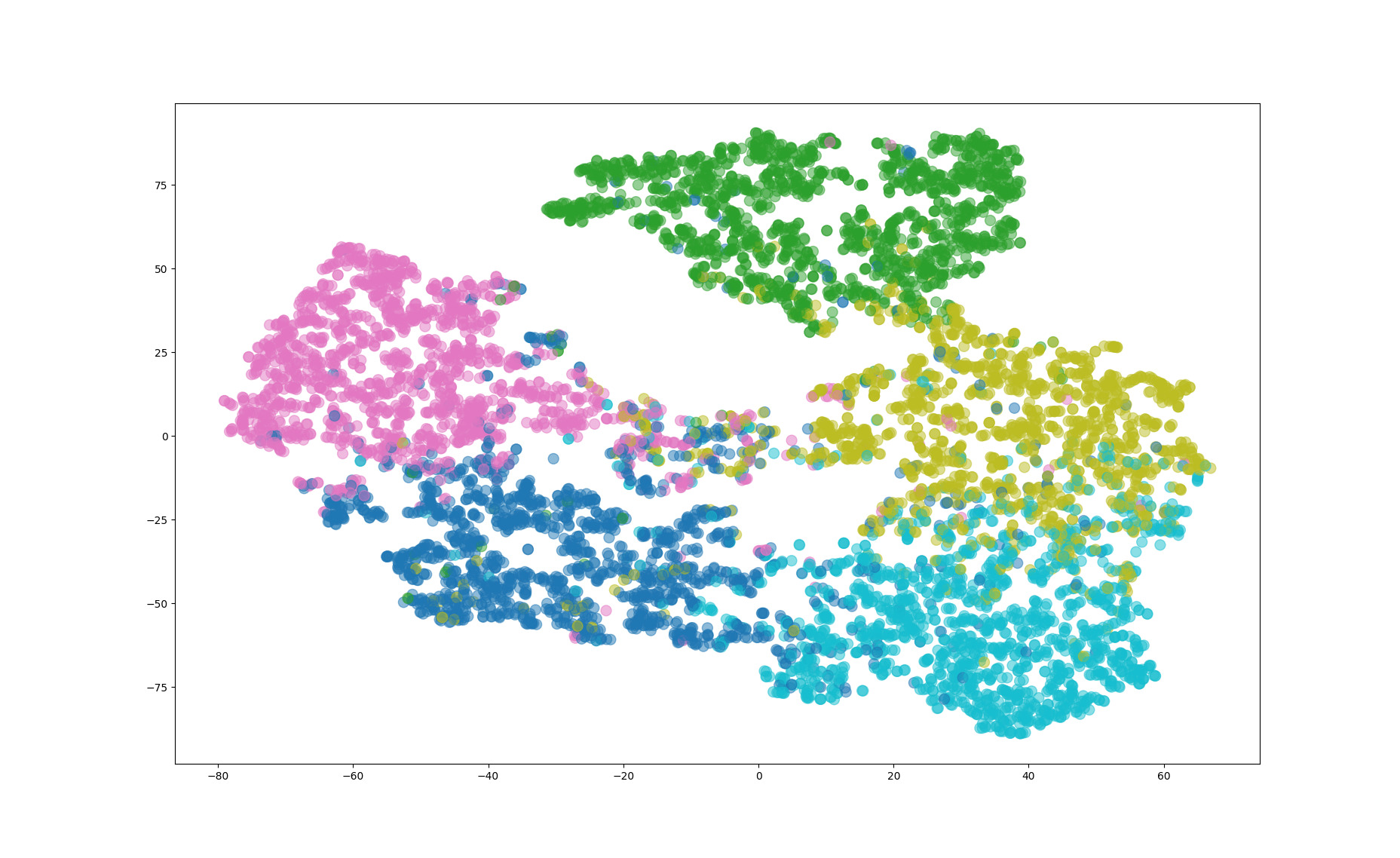}}
        \hfill
   \subfloat[(c) Epoch 30\label{e40}]{%
       \includegraphics[width=0.25\linewidth, trim={6.5cm 4cm 5cm 4cm},clip]{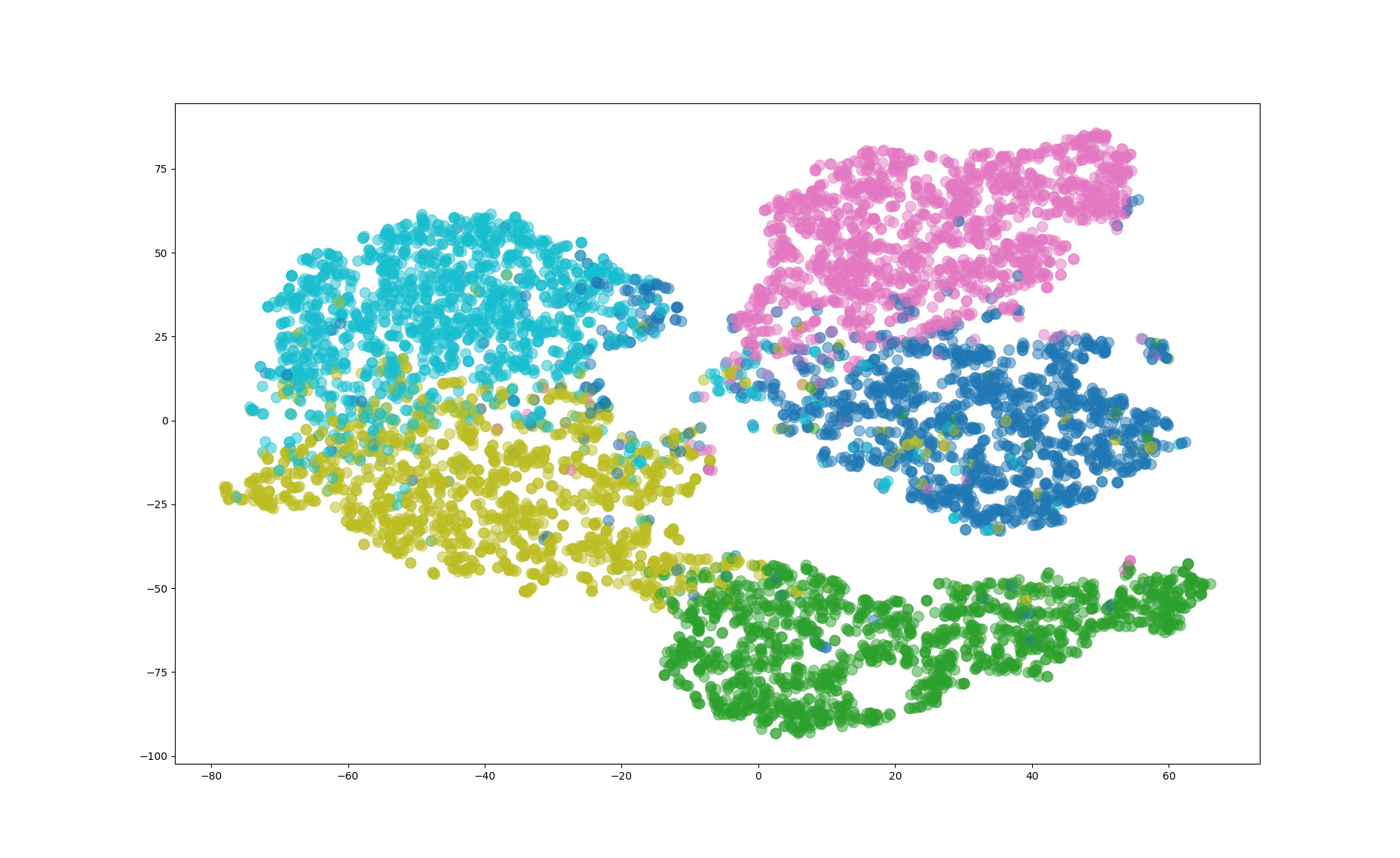}}
        \hfill
   \subfloat[(d) Epoch 90\label{e60}]{%
       \includegraphics[width=0.25\linewidth, trim={6.5cm 4cm 5cm 4cm},clip]{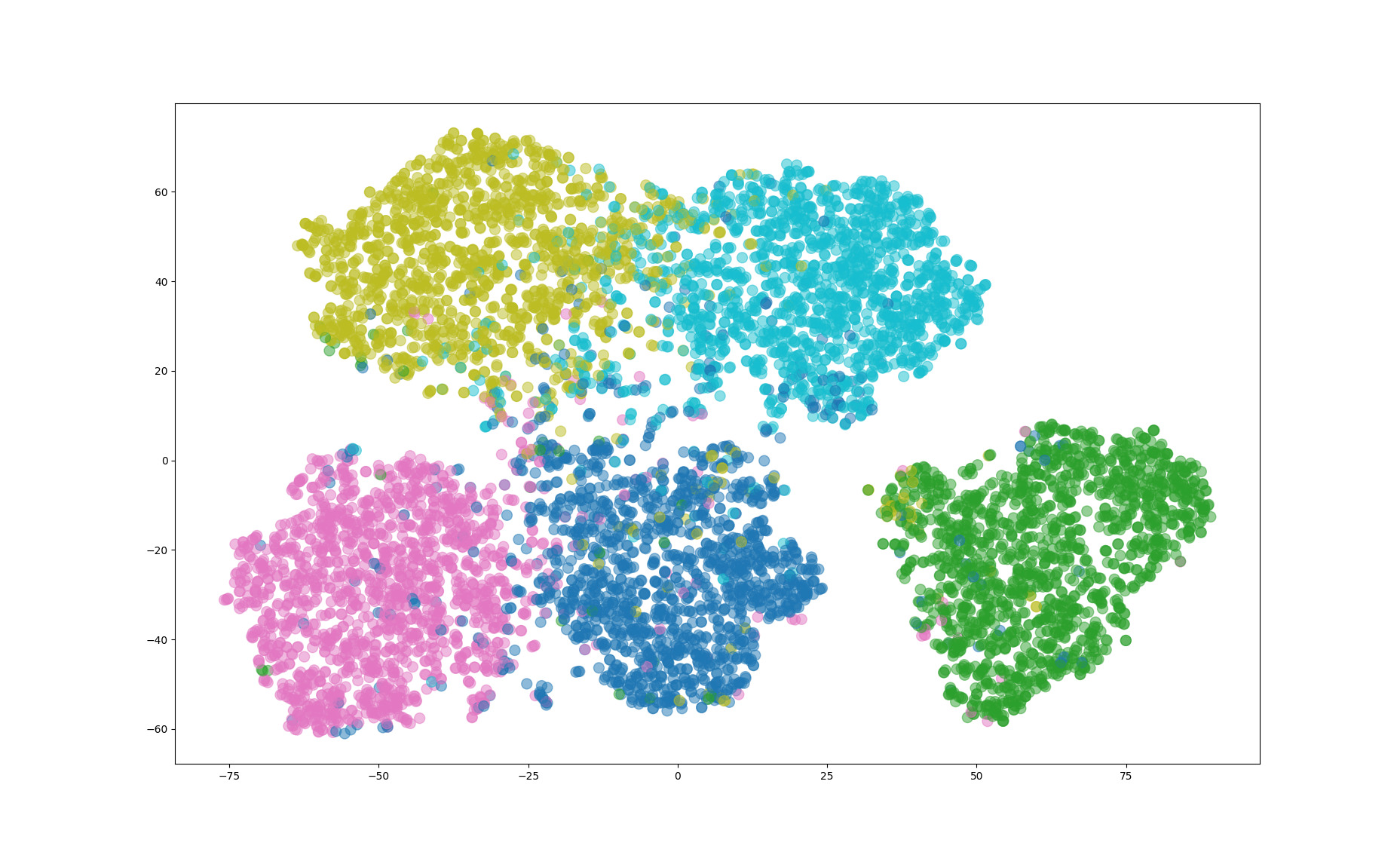}}
  \caption{Two-dimensional t-SNE visualizations of the audio samples embedded with our SE-R approach after 0, 10, 30, and 90 epochs of training. The color of the points indicate the corresponding ground truth class. We observe that clusters of embeddings are clearly separable as the training progresses and they highly correlate with the ground truth terrain class.}
  \label{fig:tsnes}
\end{figure*}

For evaluating our proposed unsupervised metric learning approach, we compare the clustering accuracy and the normalized mutual information (NMI) score of the learned embeddings on a validation split. The NMI score between two sets of clusters $Y$ and $C$ is defined as
\begin{equation}
\text{NMI}(Y,C) = \frac{2 I(Y,C)}{H(Y) + H(C)}
\end{equation}
where $I(Y,C)$ denotes the mutual information between cluster set $Y$ and $C$, and $H(Y)$ and $H(C)$ denotes the entropy of cluster sets $Y$ and $C$ respectively. We define $Y$ as the cluster set of the ground truth labels. The clustering accuracy can be obtained by assigning class labels to cluster indices that yield the highest overall accuracy such that
\begin{equation}
\text{Accuracy}(Y,C) = \frac{1}{N} \sum_k \max_j |y_k \cap c_j |
\end{equation}
where $y_k$ and $c_k$ denote particular clusters in $Y$ and $C$ respectively. Once again we select $Y$ as the cluster set of the ground truth labels. The optimal assignment in terms of clustering purity can then be obtained using the Hungarian Algorithm \cite{kuhn1955hungarian}. We use the class-wise Intersection over Union (IoU) and mean Intersection over Union (mIoU) score to quantify the performance of our self-supervised semantic segmentation model on the densely labeled validation set. However, IoU values for weakly labeled images cannot be meaningfully interpreted since only a small fraction ($\le 5 \%$) of pixels in each image are labeled. Therefore, we use the average class-wise model recall for evaluations on the weakly labeled validation set.

\subsection{Training Details}
\label{sec:network-training}

\subsubsection{Audio Clustering}

We trained our SE and SE-R models for 100 epochs using the AdaDelta optimizer with a learning rate of $\num{1e-2}$ and momentum of $0.9$. While we initially trained the comparison methods for deep embedded clustering (DEC~\cite{xie2016unsupervised}, IDEC~\cite{guo2017improved}, and DCEC~\cite{guo2017deep}) with the hyperparameters menioned in the original publications and then further optimized them on our dataset. We adopt the architecture of the audio encoder and decoder from the work of engel~\textit{et al.}~\cite{engel2017neural} and we use the MobileNet~V2~\cite{sandler2018mobilenetv2} model pre-trained on the ImageNet dataset for the visual feature extraction network that is used for forming triplets in metric learning.

\subsubsection{Self-Supervised Semantic Segmentation}

For semantic segmentation of the terrains, we adopt the AdapNet++ network architecture~\cite{valada19ijcv} with an EfficientNet backbone. We perform random augmentations during training including rotation, flipping as well as color jitter in hue, saturation, brightness, and contrast. We use the Adam optimizer with an initial learning rate of $\num{1e-3}$ and polynomial learning rate decay for training the network. We set the parameters $\beta_1 = 0.9$ and $\beta_2 = 0.999$.

\subsection{Evaluation of Unsupervised Acoustic Feature Clustering}
\label{sec:clustering-results}

\begin{table}
\centering
\caption{Comparison of NMI scores and clustering accuracies for k-means clustering of audio embeddings with different embedding models.}
\label{tab:nmi}
\begin{tabular}{p{3.4cm}cc}
\toprule
Method & NMI & Accuracy (\%) \\ 
\midrule
Image embeddings & 0.518   & 68.32 \\ 
Plain CNN Autoencoder \cite{engel2017neural}  & 0.584  & 77.89 \\ 
DEC~\citep{xie2016unsupervised} & 0.686 & 85.35 \\ 
IDEC~\citep{guo2017improved} & 0.583 & 77.56 \\
DCEC~\citep{guo2017deep} &  0.693 & 84.28 \\
\midrule
SE (Ours)  & 0.822 &  94.07 \\ 
SE-R (Ours)   & \textbf{0.839}  & \textbf{94.81}  \\ 
\bottomrule
\end{tabular} 
\end{table}

We evaluate our proposed Siamese Encoder (SE) and Siamese Encoder with Reconstruction (SE-R) models and compare them with multiple recent unsupervised deep clustering methods, namely, Deep Embedded Clustering (DEC)~\citep{xie2016unsupervised}, Improved Deep Embedded Clustering (IDEC)~\citep{guo2017improved} and Deep Clustering with Convolutional Autoencoders (DCEC)~\citep{guo2017deep}). We evaluate these models using the standard clustering metrics, NMI score and clustering accuracy. We use the same encoder-decoder architectures for all models to allow for a fair comparison and we cluster the embeddings using the k-means clustering algorithm. Results from this experiment is shown in Table~\ref{tab:nmi}. We note that our proposed method achieves the best clustering accuracy and highest NMI score. The DEC and DCEC models are able to substantially improve the clustering metrics from the plain autoencoder, however they are not able to achieve clustering accuracy over $85.35\%$. We observe that the clustering obtained from the terrain image patch embeddings yield a much lower NMI value and clustering accuracy overall. While we sample our triplets from the terrain image patch embeddings, we argue that predominantly local class neighborhood consistency is the key contributor to the success of our triplet sampling approach. With a high probability, patches of the sample class are embedded close to each other and patches from a different class are embedded further away from a given sample. We observe that using our triplet sampling strategy, $81.3\%$ of all triplets are correctly formed. For the clustering accuracies using our approach, we report values of $94.07\%$ and $94.81\%$, respectively.

\figref{fig:tsnes} shows the audio embedding clusters that we obtain using our SE-R approach, reduced to two spatial dimensions using t-SNE. We note that before training the model, there is already some correlation between the sample embedding location and the sample label visible due to the convolutional architecture of the encoder. We observe that as the training progresses, the clusters become well separable and highly correlate with the ground truth classes. The gap between clusters and the non-uniform distribution of samples in embedding space makes the clusters distinct and easily separable.

\subsection{Evaluation of Self-Supervised Semantic Segmentation}
\label{sec:sss}

\begin{table*}
\caption{Comparison of the semantic segmentation performance of our model self-supervised with SE-R, our model self-supervised with the best performing baseline model, and our model that is trained in a supervised manner with manually annotated images as ground truth. We report class-wise IoU scores for the results on the densely labeled validation set and we report the recall for the weakly labeled validation set.}
\label{tab:iou-comparison}
\centering
\setlength\tabcolsep{2pt} 
\begin{tabular}{p{2.2cm}|P{2.3cm}P{2.3cm}P{2.3cm}|P{2.3cm}P{2.3cm}P{2.3cm}}
\toprule
& \multicolumn{3}{c|}{Weakly Labeled Validation Set (Recall in $\%$)} & \multicolumn{3}{c}{Fully Labeled Validation Set (IoU in $\%$)} \\
\cmidrule{2-7}
& DEC~\citep{xie2016unsupervised} & SE-R (Ours) & & DEC~\citep{xie2016unsupervised} & SE-R (Ours) & \\
& (Self-Supervised) & (Self-Supervised) & (Supervised) & (Self-Supervised) & (Self-Supervised) & (Supervised) \\ 
\midrule
\crule[asphalt]{0.2cm}{0.2cm} Asphalt     & 0.00   & 80.05 & 86.39 & 0.07   & 67.40    & 63.47 \\  
\crule[gravel]{0.2cm}{0.2cm} Gravel      & 2.49 & 49.80 & 73.56 & 46.42 & 45.04    & 42.02 \\ 
\crule[parking]{0.2cm}{0.2cm} Parking Lot & 69.17 & 83.03 & 75.45 & 30.54 & 47.08    & 51.79 \\
\crule[grass]{0.2cm}{0.2cm} Grass        & 0.03   & 84.02 & 86.78 & 0.01   & 48.52    & 66.84 \\  
\crule[cobble]{0.2cm}{0.2cm} Cobblestone  & 42.83 & 68.97 & 66.71 & 42.45 & 62.21    & 58.76 \\ 
\midrule
\textbf{Mean}  & 22.92 & \textbf{73.21} & 77.76 & 23.93 & \textbf{54.08} & 56.51 \\ 
\bottomrule 
\end{tabular}
\end{table*}

We train the semantic segmentation network in our self-supervised framework using the weak labels that we obtain from the unsupervised acoustic feature clustering. We consider all the pixels in the image that do not have any weak labels to be of a background class which is ignored for computing the loss. We compare the evaluation metrics of the pixel-wise class predictions of a model trained on data provided by the DEC approach, our SE-R approach, and on the manually annotated data for reference. The detailed scores for each of the terrain classes in both the fully labeled evaluation validation set and the weakly labeled evaluation validation set are listed in \tabref{tab:iou-comparison}.\looseness=-1

The results demonstrate that our SE-R model provides better self-supervised generation of training data for semantic segmentation than the best performing unsupervised clustering baseline DEC. This can be attributed to the higher clustering accuracy that our model achieves which provides less noisy labels for the semantic segmentation task and thus leads to higher recall and IoU scores. Furthermore, we note that our proposed model achieves a comparable performance as the model trained in a supervised manner with manually annotated ground truth labels. The training data produced with the DEC approach contains more noisy labels due to the comparably worse audio sample embedding clusters. Our model achieves an improvement of $50.3\%$ in the recall on the weakly labeled validation set and an improvement of $30.2\%$ in the mIoU score on the fully labeled validation set, compared to the results obtained with the model trained using the DEC labels.\looseness=-1

\newcommand\items{5}   
\arrayrulecolor{white} 

\begin{figure}
\noindent
\small  
\begin{tabular}{cr*{\items}{|E}|}
\multicolumn{1}{c}{} &\multicolumn{1}{c}{} &\multicolumn{\items}{c}{Predicted} \\ \hhline{~*\items{|-}|}
\multicolumn{1}{c}{} & 
\multicolumn{1}{c}{} & 
\multicolumn{1}{c}{\rot{\enspace Parking Lot}} & 
\multicolumn{1}{c}{\rot{\enspace Grass}} & 
\multicolumn{1}{c}{\rot{\enspace Gravel}} & 
\multicolumn{1}{c}{\rot{\enspace Asphalt}} & 
\multicolumn{1}{c}{\rot{\enspace Cobblestone}} \\ \hhline{~*\items{|-}|}
\multirow{\items}{*}{\rotatebox{90}{Actual}} 
& Parking Lot       & 85.56 & 6.54 & 3.96 & 1.61 & 2.32  \\ \hhline{~*\items{|-}|}
& Grass             & 5.70 & 81.11 & 9.35 & 3.79 & 0.00  \\ \hhline{~*\items{|-}|}
& Gravel            & 3.67 & 27.66 & 54.18 & 8.06 & 6.43 \\ \hhline{~*\items{|-}|}
& Asphalt           & 2.32 & 8.86 & 6.76 & 84.34 & 5.69  \\ \hhline{~*\items{|-}|}
& Cobblestone       & 16.04 & 2.09 & 1.48 & 9.45 & 73.01 \\ \hhline{~*\items{|-}|}
\end{tabular}
  \caption{Normalized confusion matrix (in $\%$) of our self-supervised semantic terrain segmentation model trained using SE-R labels. Results are shown on the fully labeled validation set.}
  \label{fig:confusion} 
\end{figure}

\arrayrulecolor{black} 

\figref{fig:confusion} shows the confusion matrix for our self-supervised semantic terrain segmentation model trained using the SE-R labels. The results are shown on the fully labeled validation set. We observe that the \textit{Gravel} class is occasionally confused with the \textit{Grass} class. This misclassification can be attributed to the  fact that many gravel path patches are partially overgrown with grass which cannot be accounted for when manually labeling the images. We also note that some pixels predicted as the \textit{Cobblestone} class are confused with the terrain \textit{Parking Lot} class which can be attributed to the similar rectangular terrain patterns in both classes.

\subsection{Qualitative Semantic Terrain Segmentation Results}
\label{sec:qualitative}

\begin{figure*}
\centering
\footnotesize
\setlength{\tabcolsep}{0.1cm}
{\renewcommand{\arraystretch}{2}
    \begin{tabular}{P{0.5cm}P{4.1cm}P{4.1cm}P{4.1cm}P{4.1cm}}
    \arrayrulecolor{red}
        & Input Image & Output (DEC~\citep{xie2016unsupervised} Labels) & Output (SE-R (Ours) Labels) & Improvement\textbackslash Error Map \\
        \X{(a)} & \includegraphics[width=\linewidth,trim={2cm 3cm 2cm 3cm},clip]{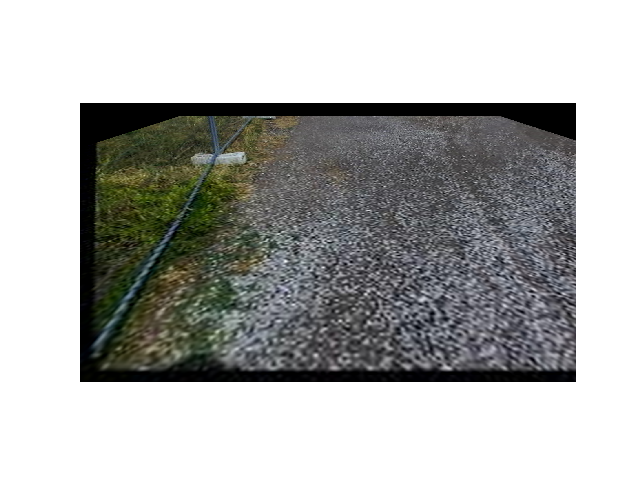} & \includegraphics[width=\linewidth,trim={2cm 3cm 2cm 3cm},clip]{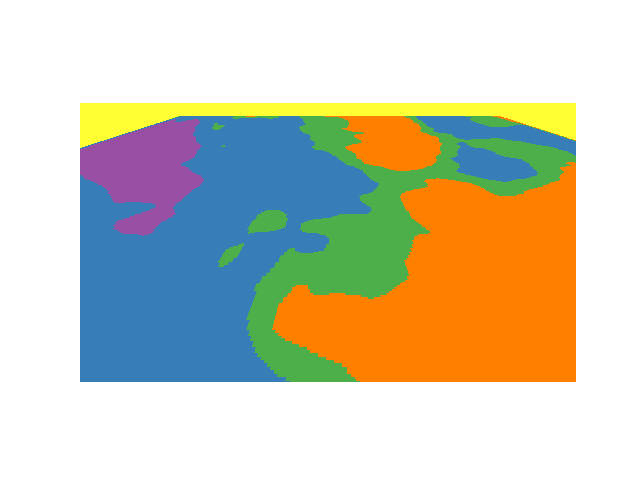} & \includegraphics[width=\linewidth,trim={2cm 3cm 2cm 3cm},clip]{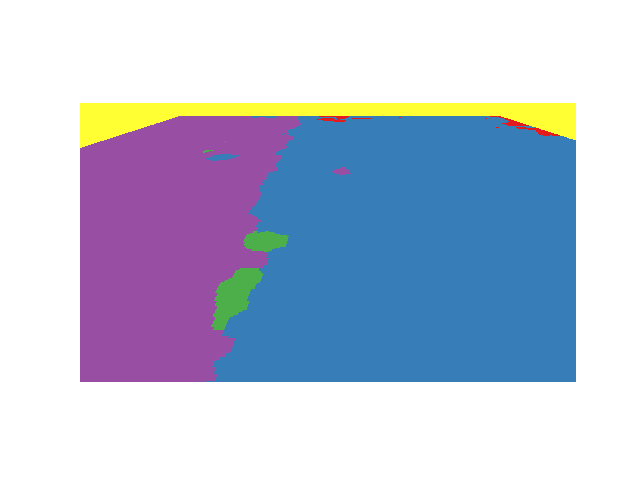} & \includegraphics[width=\linewidth,trim={2cm 3cm 2cm 3cm},clip]{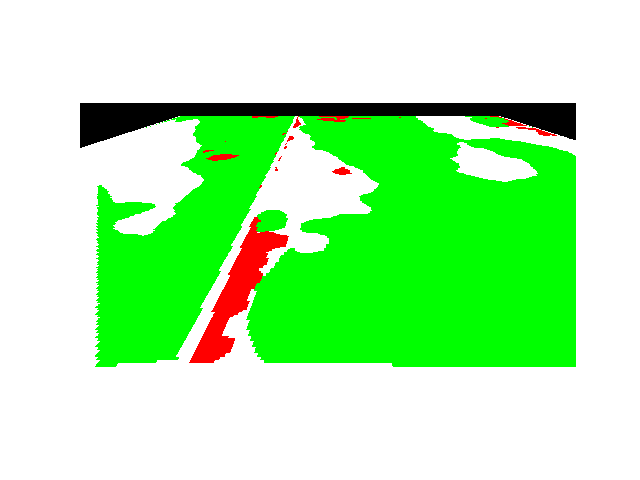} \\
        \X{(b)} & \includegraphics[width=\linewidth,trim={2cm 3cm 2cm 3cm},clip]{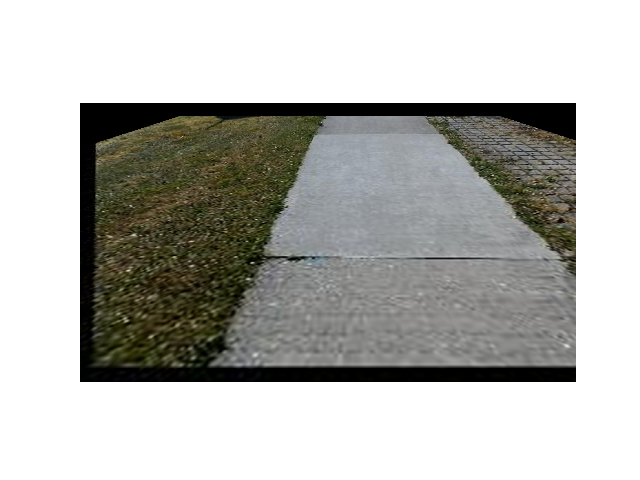} & \includegraphics[width=\linewidth,trim={2cm 3cm 2cm 3cm},clip]{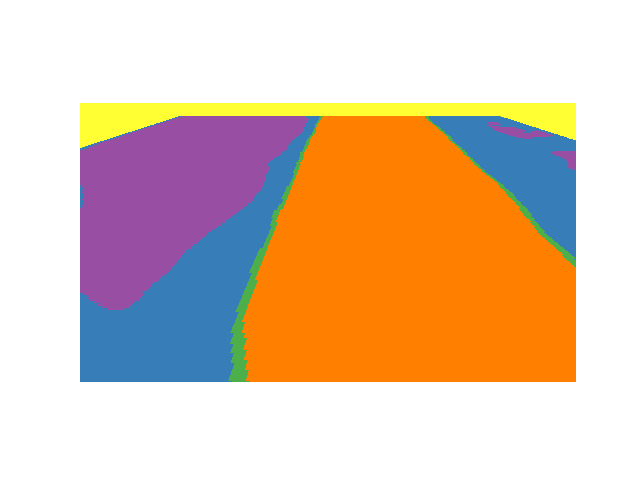} & \includegraphics[width=\linewidth,trim={2cm 3cm 2cm 3cm},clip]{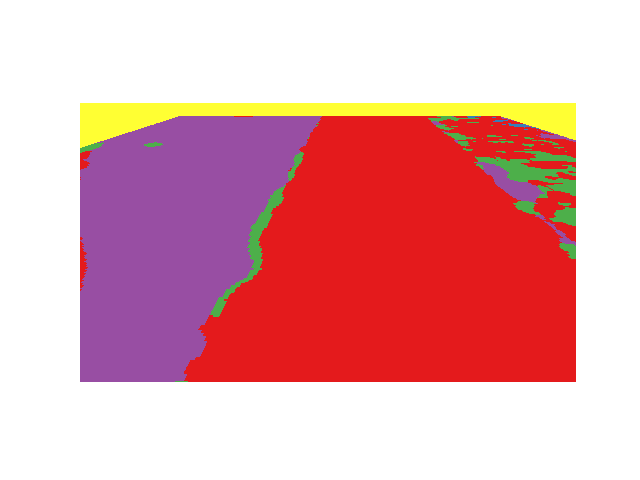} & \includegraphics[width=\linewidth,trim={2cm 3cm 2cm 3cm},clip]{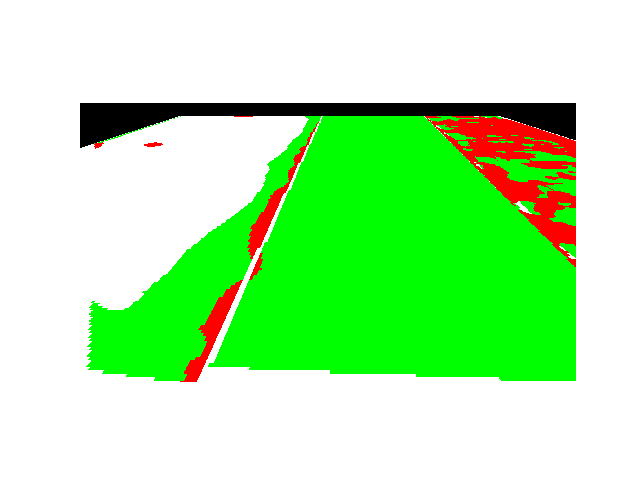} \\
        \X{(c)} & \includegraphics[width=\linewidth,trim={2cm 3cm 2cm 3cm},clip]{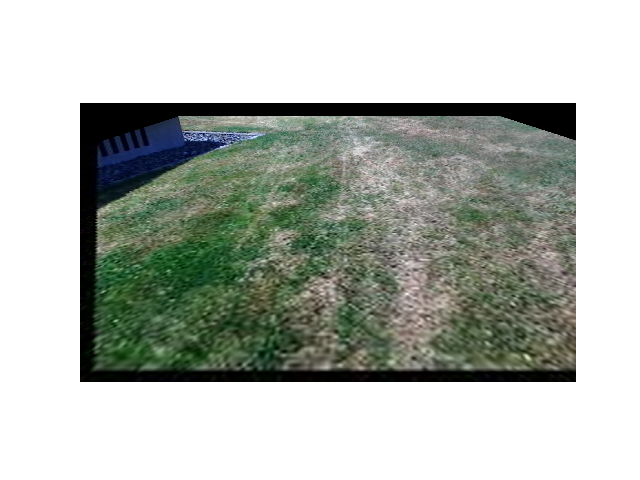} & \includegraphics[width=\linewidth,trim={2cm 3cm 2cm 3cm},clip]{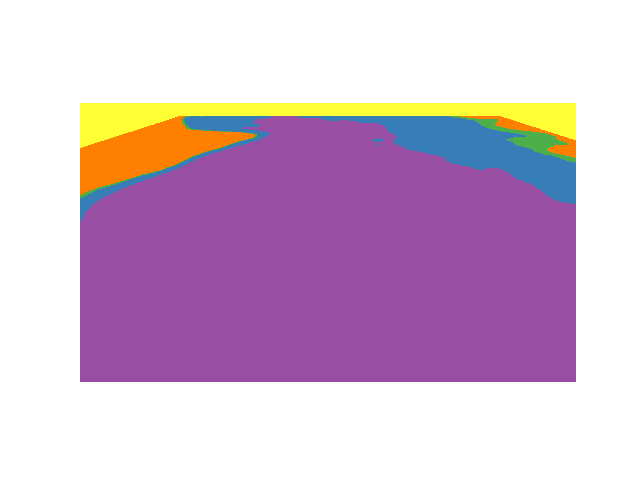} & \includegraphics[width=\linewidth,trim={2cm 3cm 2cm 3cm},clip]{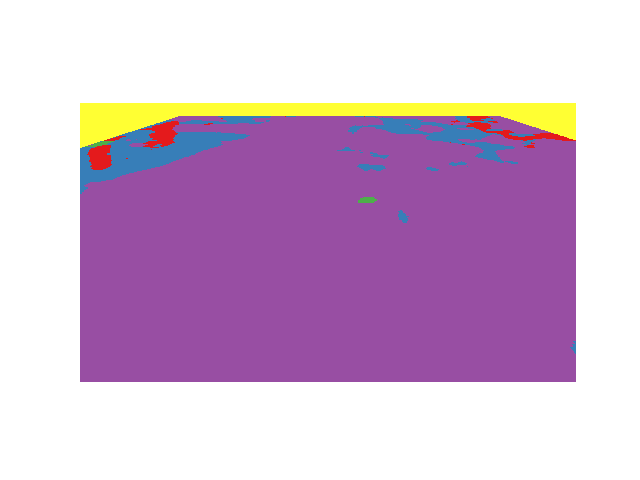} & \includegraphics[width=\linewidth,trim={2cm 3cm 2cm 3cm},clip]{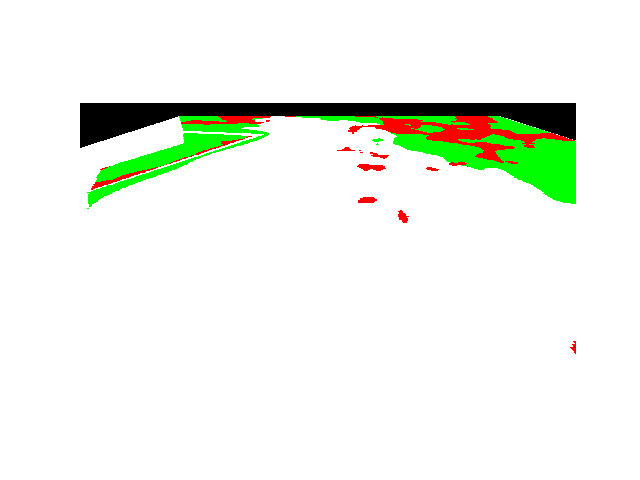} \\
        \X{(d)} & \includegraphics[width=\linewidth,trim={2cm 3cm 2cm 3cm},clip]{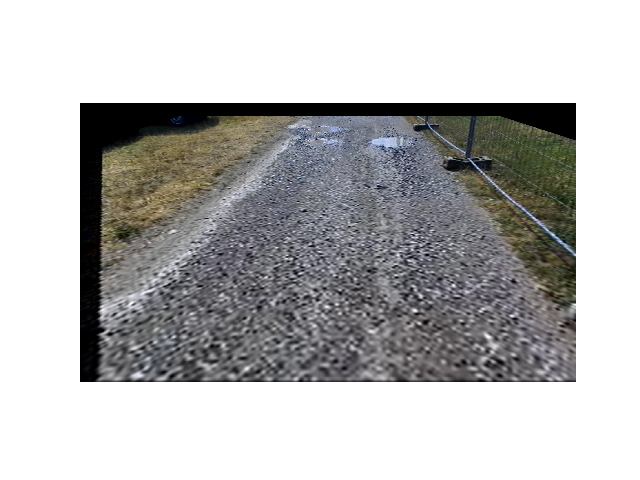} & \includegraphics[width=\linewidth,trim={2cm 3cm 2cm 3cm},clip]{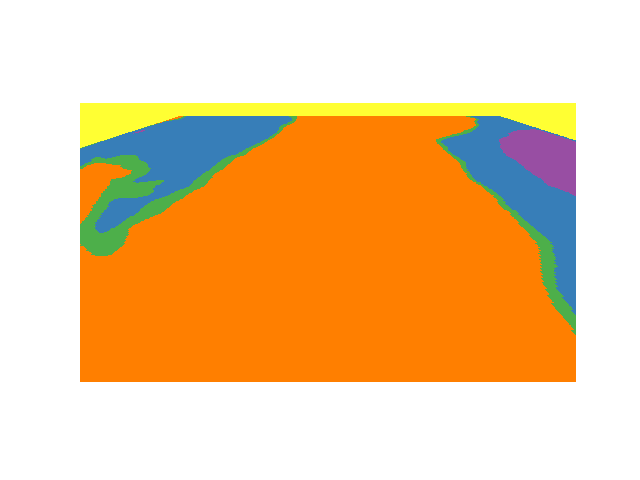} & \includegraphics[width=\linewidth,trim={2cm 3cm 2cm 3cm},clip]{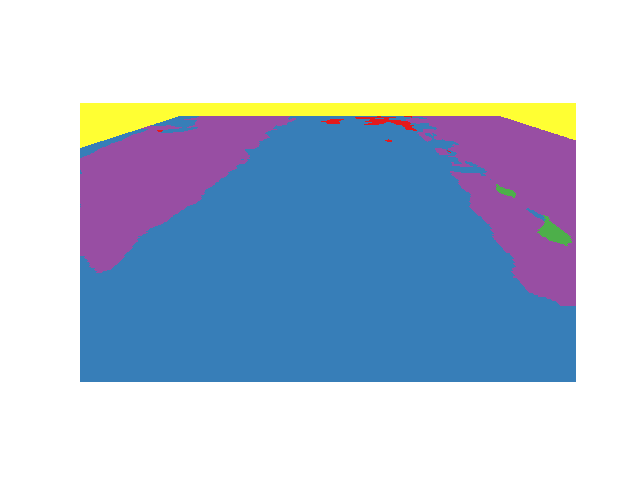} & \includegraphics[width=\linewidth,trim={2cm 3cm 2cm 3cm},clip]{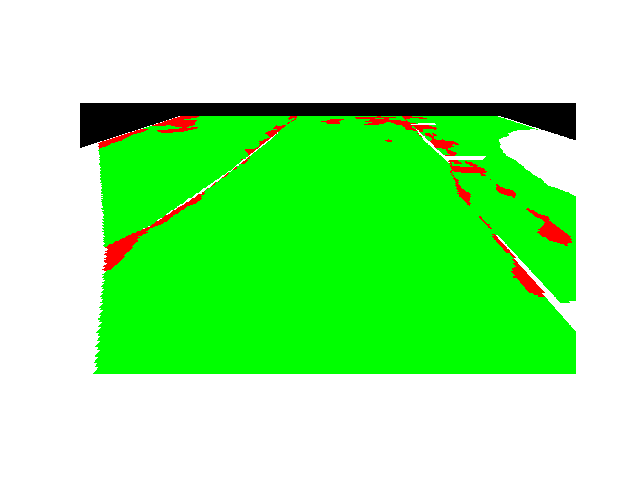} \\
        \X{(e)} & \includegraphics[width=\linewidth,trim={2cm 3cm 2cm 3cm},clip]{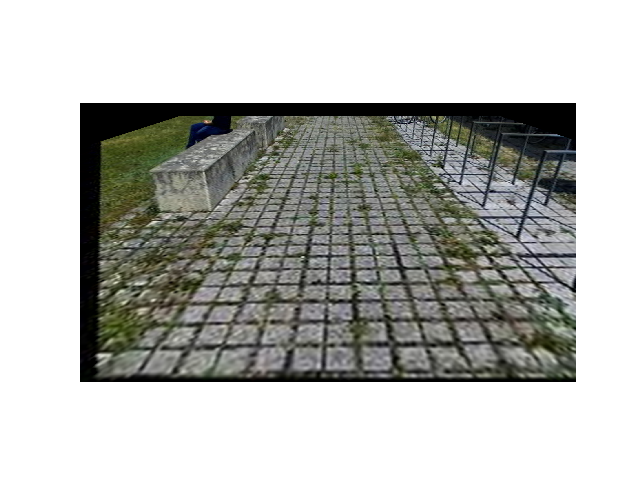} & \includegraphics[width=\linewidth,trim={2cm 3cm 2cm 3cm},clip]{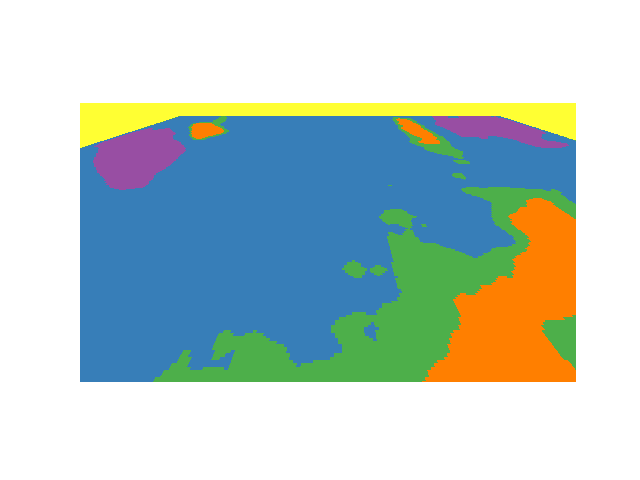} & \includegraphics[width=\linewidth,trim={2cm 3cm 2cm 3cm},clip]{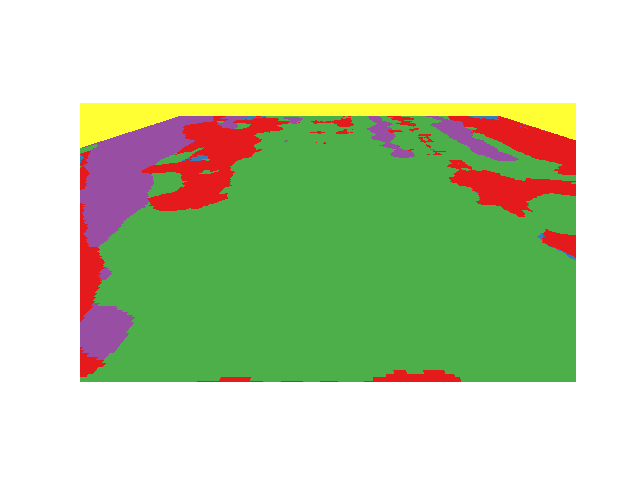} & \includegraphics[width=\linewidth,trim={2cm 3cm 2cm 3cm},clip]{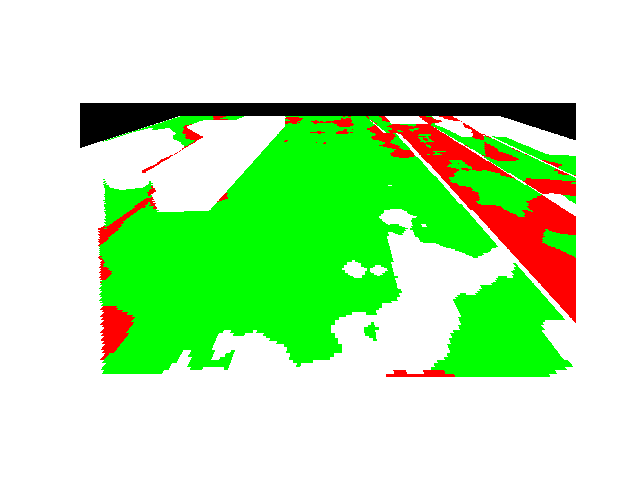} \\
        \X{(f)} & \includegraphics[width=\linewidth,trim={2cm 3cm 2cm 3cm},clip]{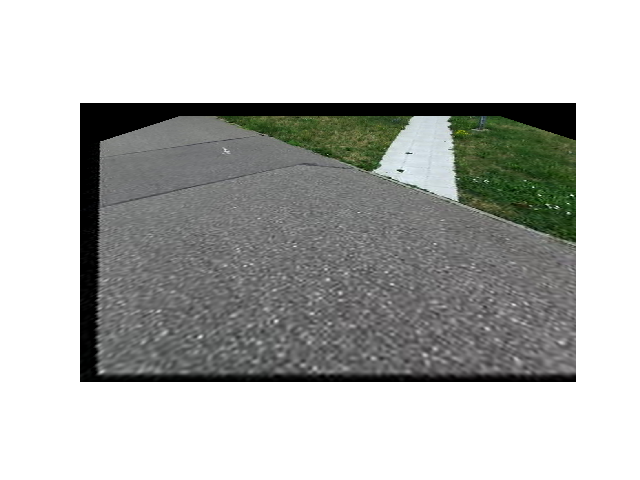} & \includegraphics[width=\linewidth,trim={2cm 3cm 2cm 3cm},clip]{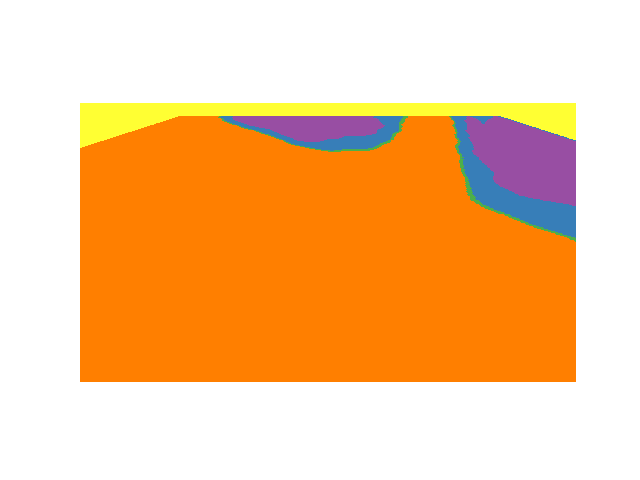} & \includegraphics[width=\linewidth,trim={2cm 3cm 2cm 3cm},clip]{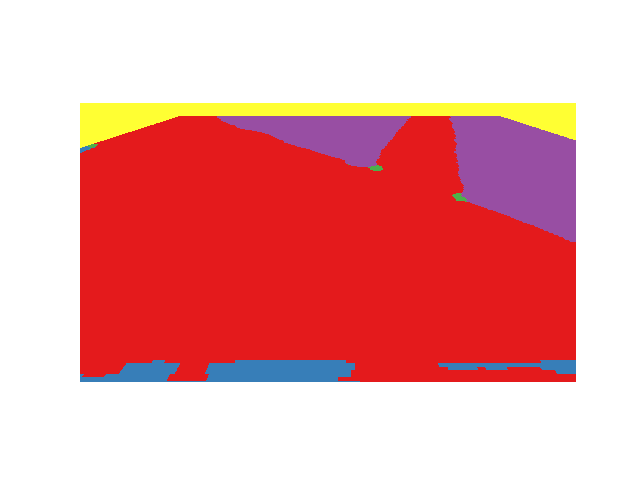} & \includegraphics[width=\linewidth,trim={2cm 3cm 2cm 3cm},clip]{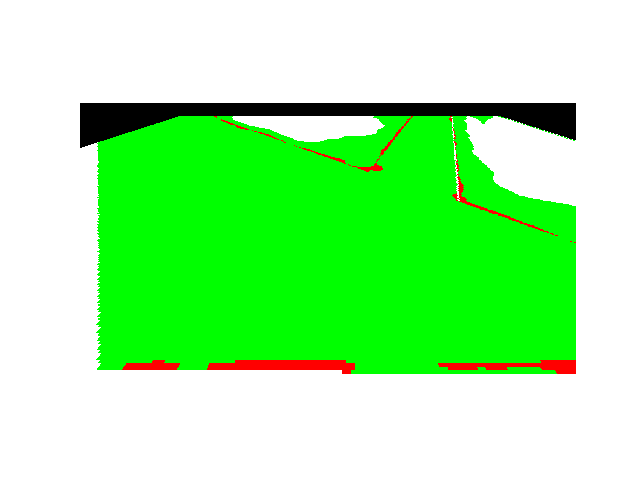} \\
        \X{(g)} & \includegraphics[width=\linewidth,trim={2cm 3cm 2cm 3cm},clip]{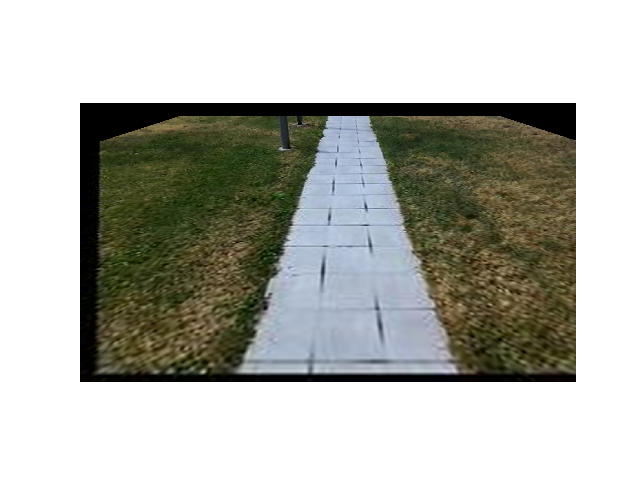} & \includegraphics[width=\linewidth,trim={2cm 3cm 2cm 3cm},clip]{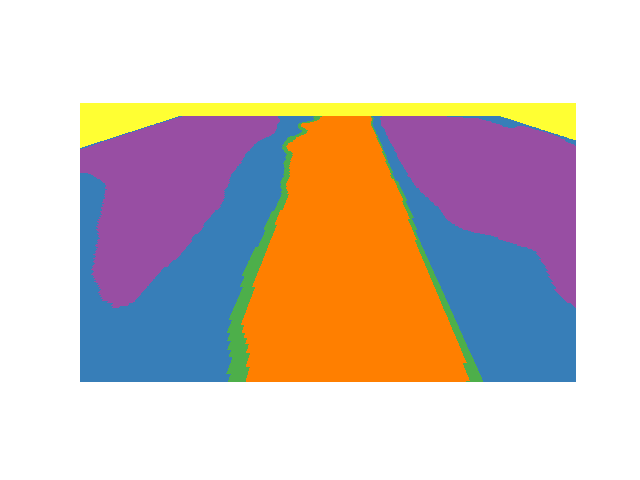} & \includegraphics[width=\linewidth,trim={2cm 3cm 2cm 3cm},clip]{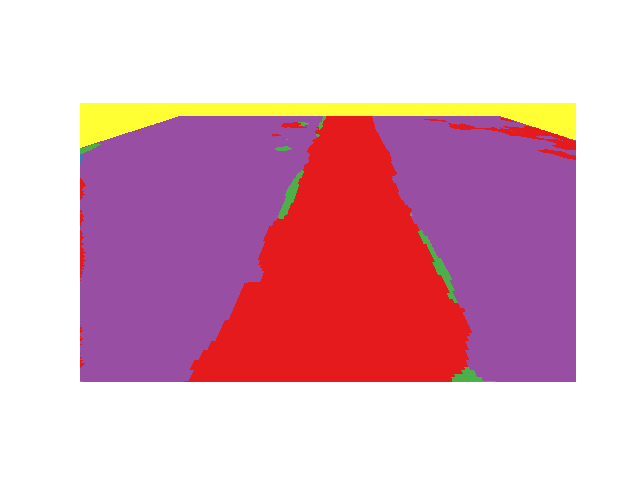} & \includegraphics[width=\linewidth,trim={2cm 3cm 2cm 3cm},clip]{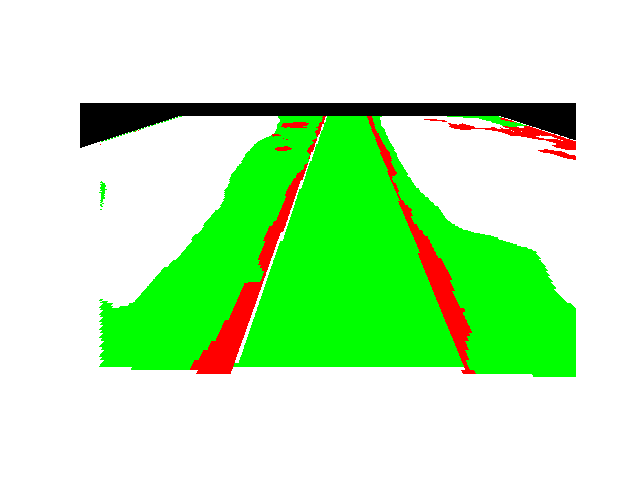} \\
        \midrule[0.05cm]
        \X{(h)} & \includegraphics[width=\linewidth,trim={2cm 3cm 2cm 3cm},clip]{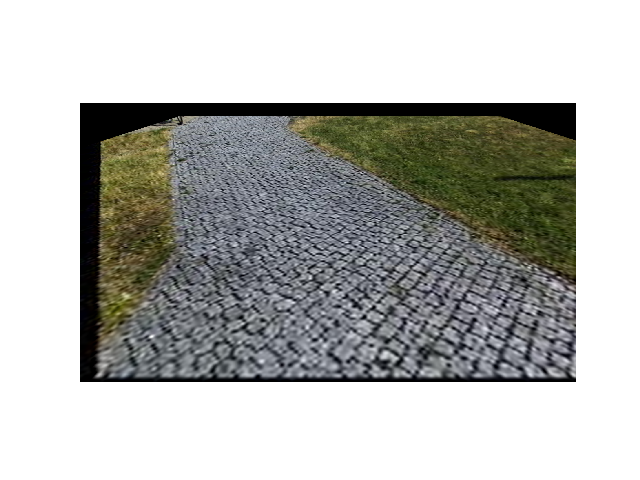} & \includegraphics[width=\linewidth,trim={2cm 3cm 2cm 3cm},clip]{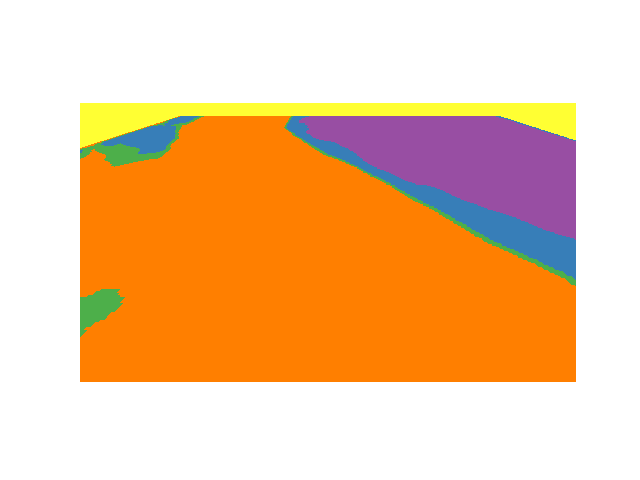} & \includegraphics[width=\linewidth,trim={2cm 3cm 2cm 3cm},clip]{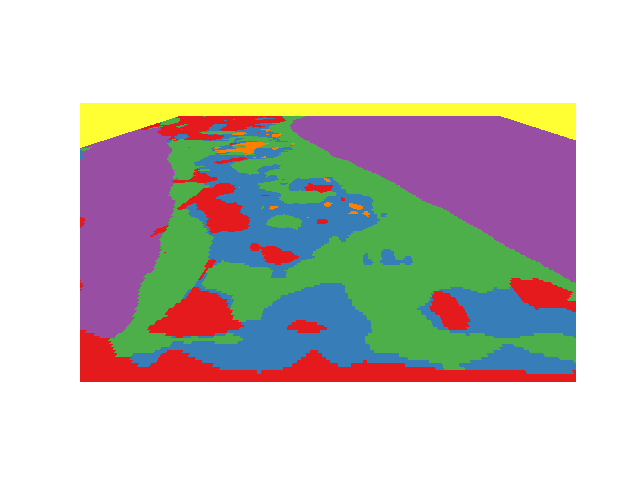} & \includegraphics[width=\linewidth,trim={2cm 3cm 2cm 3cm},clip]{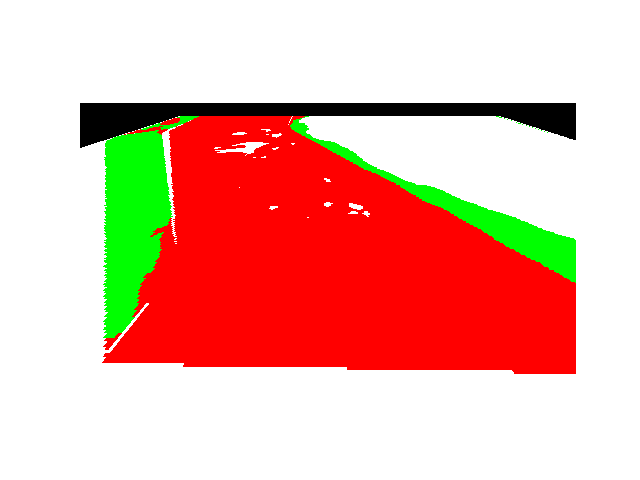} \\
        \X{(i)} & \includegraphics[width=\linewidth,trim={2cm 3cm 2cm 3cm},clip]{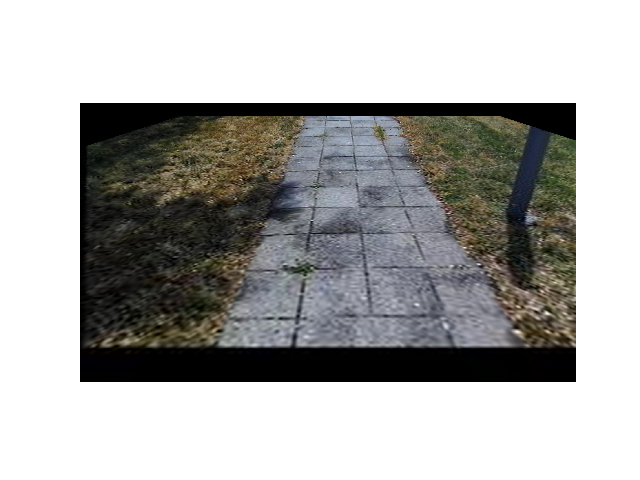} & \includegraphics[width=\linewidth,trim={2cm 3cm 2cm 3cm},clip]{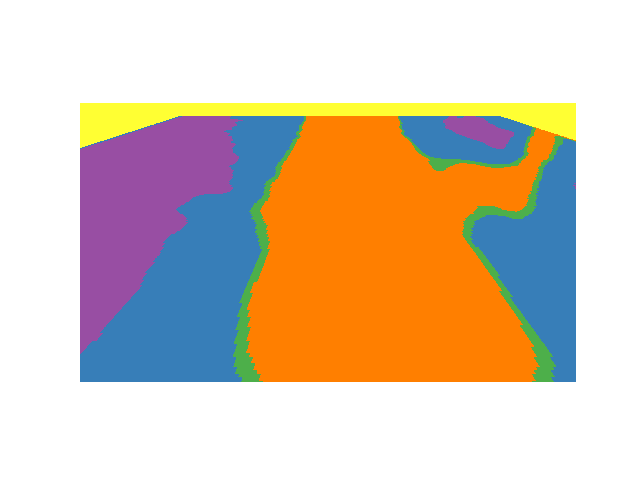} & \includegraphics[width=\linewidth,trim={2cm 3cm 2cm 3cm},clip]{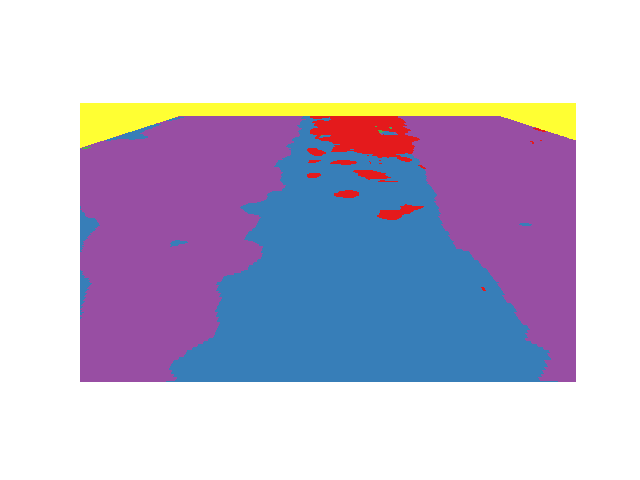} & \includegraphics[width=\linewidth,trim={2cm 3cm 2cm 3cm},clip]{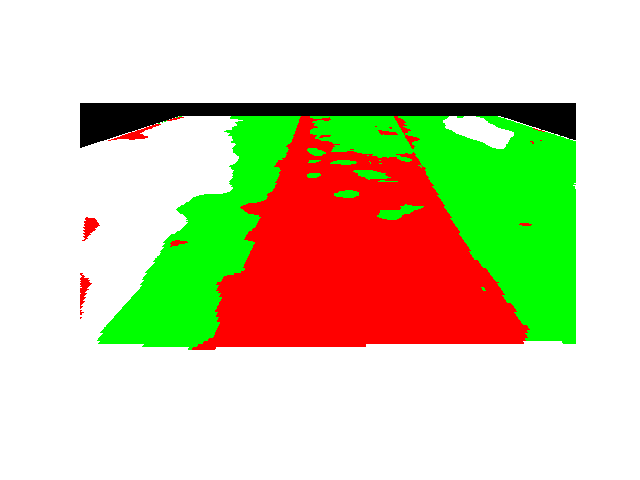}
    \end{tabular}}
    \caption{Qualitative semantic segmentation results of our model self-supervised by our unsupervised SE-R approach in comparison to the model self-supervised using the unsupervised DEC method. We also show the improvement/error map which depicts the misclassified pixels in red and the pixels that are misclassified by the segmentation model trained on labels from DEC but correctly predicted by SE-R in green. The last two rows show failure modes of our model. The legend for terrain labels correspond to the colors shown in \tabref{tab:iou-comparison}.}
    \label{fig:predictions} 
\end{figure*}

In this section, we qualitatively evaluate the performance of our proposed semantic terrain segmentation model trained on SE-R labels with the model trained on the DEC labels. \figref{fig:predictions} shows the input image transformed into the original perspective along with the semantic segmentation outputs and the Improvement\textbackslash Error Map which shows the improvement achieved by training on our SE-R labels over training on DEC labels, in green and the pixels that are misclassified by our model trained on SE-R labels in comparison to the ground truth in red.\looseness=-1

We observe that for the majority of the scenes, the network trained with our self-supervised SE-R approach is able to predict the semantic classes accurately. Our model is also able to segment complex scenes with visual clutter and multiple terrains bleeding into one and another. We can see that some misclassifications are produced due to high-contrast shadows and visual ambiguities (i.e. classes such as \textit{Asphalt} and \textit{Gravel}). In contrast, the segmentation network trained on the labels obtained with the DEC model fails to predict the correct semantic class in a significant number of scenes. While it is able to mostly predict the borders between different terrain classes due to the change in color and texture, it fails to find the correct correspondence between terrain appearance and underlying terrain class. This can be attributed to the inconsistencies in labeling the images with the labels provided by the DEC model. From the Improvement\textbackslash Error Map, we can observe that the segmentation network trained on labels from our SE-R approach improves substantially over the output of the baseline labeling approach. \figref{fig:predictions}~(h) and (i) show failure cases of our approach where the asphalt road and cobblestone are misclassified due to the high visual similarity to other gravel areas in the training dataset. This problem can be alleviated be retraversing such regions and fine-tuning our model on the new samples as we show in the results in \figref{fig:compare-models}.\looseness=-1

\subsection{Ablation studies}
\label{sec:ablation}

In this section, we present results from several ablation studies that we performed to evaluate and analyze the performance of our model under various settings. We also present results that show the improvement in performance due to the contributions that we make in this work.

\subsubsection{Noise Resistance of Audio Embeddings}
\label{noiseresistance}

\begin{figure}
\begin{tikzpicture}
  \centering
	\begin{axis}[
		height=5cm,
		width=8.4cm,
		grid=both,
		label style={font=\footnotesize},
		tick label style={font=\footnotesize},
		legend style={font=\footnotesize},
		legend pos=north west,
		xlabel={Signal-to-Noise-Ratio [dB]},
    	ylabel={Clustering Accuracy ($\%$)},
		table/col sep=tab,
		cycle list name=exotic,
	]
	\addplot table[x=SNR,y=SAE] {results/snr.csv};
		\addlegendentry{SE}
	\addplot table[x=SNR,y=SAER] {results/snr.csv};
		\addlegendentry{SE-R}
	\end{axis}
\end{tikzpicture}
  \caption{Clustering accuracy of our two proposed Siamese Encoder variants for validation audio samples that are corrupted with varying amount of noise. Our SE-R approach achieves a higher clustering accuracy than the SE variant regardless of the Signal-to-Noise-Ratio.}
  \label{fig:noise} 
\end{figure}
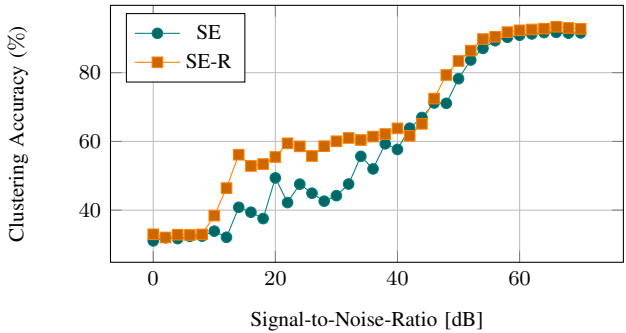

Disruptive ambient sounds such as noise of the wind, construction sounds or audio from people talking will often be present in most urban environment and it will likely be captured by the microphone mounted on the robot. This external environmental noise can influence the position of the audio sample embeddings and thus can in turn corrupt the clustering accuracy. Our approach relies on good clustering accuracy in order to accurately self-supervise the labeling of terrain images and it is therefore important for the approach to be robust to noise. We evaluate the noise robustness of the SE and SE-R variants of our model in terms of the clustering accuracy with varying amount of white noise added to the audio samples in our dataset. We sample the noise from a Gaussian distribution with zero mean and a standard deviation that is defined by the specific SNR. We then train our networks on the original training dataset and validate it on a disjoint noisy dataset. \figref{fig:noise} shows the dependency of the clustering validation accuracy on the signal-to-noise-ratio (SNR). We observe that the clustering accuracy drops noticeably for SNR values below $\SI{50}{\decibel}$ with both the Siamese Encoder variants and SNR values below $\SI{10}{\decibel}$ yield accuracies no better than chance. Nevertheless, our SE-R variant maintains a higher accuracy than the SE variant for almost the whole range of SNR values which shows the superior noise resilience of our SE-R model.

\subsubsection{Number of Samples for Clustering}
\label{datastudy}

\begin{figure}
\begin{tikzpicture}
  \centering

	\begin{semilogxaxis}[
		height=5cm,
		width=8cm,
		grid=both,
		ymax=100.0,
		ymin=40.0,
		label style={font=\footnotesize},
		tick label style={font=\footnotesize},
		legend style={font=\footnotesize},
		legend pos=north west,
		xlabel={Number of Triplets},
    	ylabel={Clustering Accuracy ($\%$)},
		table/col sep=tab,
		cycle list name=exotic,
	]

    \addplot coordinates {
		(30, 47.46)
		(90, 58.87)
		(150, 51.59)
		(300, 59.72)
		(900, 68.20)
		(1500, 71.20)
		(3000, 80.50)
		(9000, 92.10)
		(18000, 93.9)
	};
	\addlegendentry{SE}

	\addplot coordinates {
		(30, 48.64)
		(90, 56.11)
		(150, 54.34)
		(300, 60.88)
		(900, 66.60)
		(1500, 72.00)
		(3000, 85.70)
		(9000, 92.8)
		(18000, 94.8)
	};
	
	\addlegendentry{SE-R}

	\end{semilogxaxis}
\end{tikzpicture}
  \caption{Clustering accuracy of our proposed SE and SE-R models when trained on different numbers of triplets. The clustering accuracy increases with the number of triplets used for training. }
  \label{fig:n_samples} 
\end{figure}
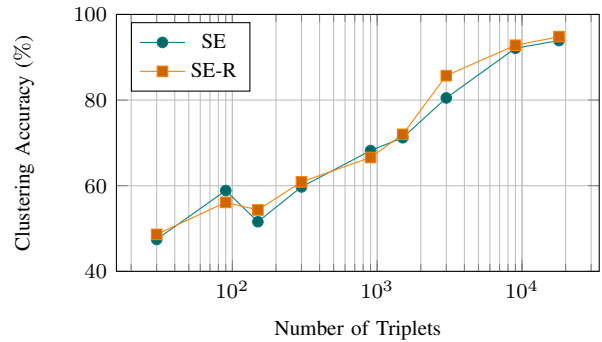

Deep learning methods require a significant amount of training data and they generally yield a better performance with increasing amounts of the right training data. In order to study the influence of this factor on the proposed framework, we investigate the dependency of the clustering accuracy on the number of triplets used for training. \figref{fig:n_samples} shows results from this experiment where it can be seen that the clustering accuracy benefits from large number of triplets. This can be attributed to the fact that the encoder can generalize effectively to unseen samples when it is trained on a large number of samples. 

\subsubsection{Triplet Sampling Mechanism}
\label{tripletstudy}

\begin{table}
\caption{Influence of the triplet sampling method on audio clustering accuracy of our SE-R approach. Correct triplets and clustering accuracy in $\%$.}
\label{triplet-study-table}
\centering
\setlength\tabcolsep{2pt} 
\begin{tabular}{p{2.9cm} c c c}
\toprule
Sampling Mechanism &  Correct Triplets & Clustering Accuracy & NMI \\ \midrule
Ground truth & 100.0 & 98.48 & 0.945 \\ \midrule
Random & 16.05  & 32.10 & 0.259 \\  
\makecell[l]{$x^+$: cluster-based \\ $x^-$: distance-based} & 49.23  & 62.45 & 0.433 \\ 
\makecell[l]{$x^+$: cluster-based \\ $x^-$: cluster-based} & 54.04  & 82.75  & 0.670 \\ 
\makecell[l]{$x^+$: distance-based \\ $x^-$: distance-based}  & 75.51  & 65.05  & 0.488 \\  
\makecell[l]{$x^+$: distance-based \\ $x^-$: cluster-based} & \textbf{81.35}  & \textbf{94.80}  & \textbf{0.839} \\
\bottomrule
\end{tabular} 
\end{table}

In order to analyze the influence of the triplet sampling mechanism on the clustering accuracy of the audio samples, we performed experiments by investigating different ways of sampling triplets from the feature space of the terrain images. For each anchor sample $x^0$, we find a positive sample $x^+$ and a negative sample $x^-$ according to the following heuristics:
\begin{itemize}
    \item \textbf{Random}: $x^+$ and $x^-$ are randomly selected from the training dataset.
    \item \textbf{Distance-based}: We first randomly select anchor samples. Then we select the closest sample in visual embedding space as a positive sample and the most distant sample as a negative sample.
    \item \textbf{Cluster-based}: We first cluster the visual embedding space using k-means clustering. We then randomly select anchor samples and a random sample within the same cluster as the anchor sample, as a positive example. Subsequently, we select a random sample within a different cluster as a negative sample.
    \item \textbf{Ground truth}: We formulate triplets according to their known ground truth terrain class. This serves as a reference.
\end{itemize}

Note that of all aforementioned heuristics, only the ground truth based triplet forming requires knowledge about the class of the samples. \tabref{triplet-study-table} shows the percentage of correctly formed triplets, the resulting clustering accuracy, and NMI using our SE-R approach for each triplet forming mechanism. In addition to purely cluster-based or distance-based triplet forming, we also evaluate combinations of the two methods. We observe that random triplet forming leads to a low number of correctly formed triplets and a low clustering accuracy that is marginally above chance. The cluster-based and distance-based triplet forming mechanisms yield varying fraction of correctly formed triplets, where the mechanism with $x^+$ distance-based and $x^-$ cluster-based yields triplets of which $81.35\%$ are formed correctly. Training our SE-R network on triplets generated with this mechanism also yields the reported highest clustering accuracy of $94.8\%$. From these results we can infer that a low fraction of correctly formed triplets leads to a lower clustering accuracy and NMI. This can be attributed to the fact that a higher number of correctly formed triplets outweighs the few incorrectly formed triplets and still forces the respective audio samples to be embedded into the right location in embedding space. For a reference, we also report the clustering accuracy and NMI for triplets which were formed using the ground truth class of the audio samples which we do not assume that we have for the other sampling methods as this would make the approach supervised. 

\subsubsection{Ratio of Correctly Formed Triplets}

\begin{figure}
\begin{tikzpicture}
  \centering

	\begin{axis}[
		height=5cm,
		width=8cm,
		grid=both,
		ymax=100.0,
		ymin=0.0,
		label style={font=\footnotesize},
		tick label style={font=\footnotesize},
		legend style={font=\footnotesize},
		legend pos=north west,
		xlabel={Ratio of Correctly Formed Triplets},
    	ylabel={Clustering Accuracy ($\%$)},
		table/col sep=tab,
		cycle list name=exotic,
	]

	\addplot coordinates {
		(0.0, 25)
		(0.1, 27)
		(0.2, 25)
		(0.3, 32)
    	(0.4, 50)	
    	(0.5, 69)	
    	(0.6, 86)
    	(0.7, 87)
    	(0.8, 88)
    	(0.9, 93)
    	(1.0, 95)
	};
	
	\addlegendentry{SE}

	\addplot coordinates {
		(0.0, 32.0)
		(0.1, 30.3)
		(0.2, 32.3)
		(0.3, 38.8)
    	(0.4, 58.2)	
    	(0.5, 78.5)	
    	(0.6, 88.9)
    	(0.7, 89.2)
    	(0.8, 93.3)
    	(0.9, 95.1)
    	(1.0, 98.5)
	};
	
	\addlegendentry{SE-R}

	\end{axis}
\end{tikzpicture}
  \caption{Influence of the ratio of correctly formed triplets on the clustering accuracy for our SE-R and SE models. The clustering accuracy dramatically increases with the fraction of correctly formed triplets for both models.}
  \label{fig:fraction_triplets} 
\end{figure}
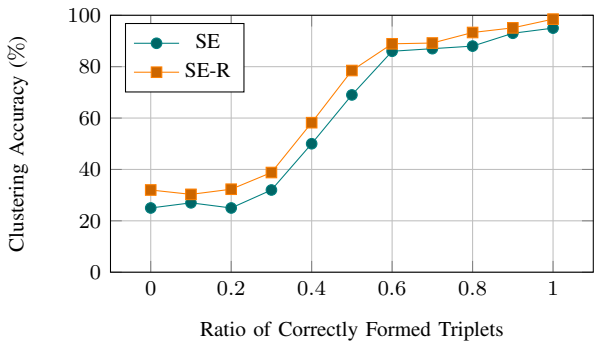

We also investigated the concrete dependency between the fraction of correctly assigned triplets and the final clustering accuracy by deliberately assigning incorrect samples as positive or negative samples, respectively. \figref{fig:fraction_triplets} shows how the clustering accuracy changes with the ratio of correctly formed triplets. We observe that the clustering accuracy increases with the fraction of correctly formed triplets. The clustering accuracy drops below $90\%$ using our SE-R approach with a correct triplet ratio of $0.8$ and monotonically decreases to values of 30\% for correct triplet ratios lower than $0.2$. We remark that this performance is consistent with the observations made in the previous section. We also observe that our SE-R variant consistently yields higher clustering accuracies than our SE variant.

\subsubsection{Segmentation Network Architectures}
\label{ablationstudy}

\begin{table}
\centering
\caption{Comparision of the self-supervised semantic terrain segmentation performance with different network architectures.}
\label{tab:ablation}
\begin{tabular}{p{2.2cm}p{1.7cm}ccc}
\toprule
Model & Backbone & Params & mIoU & Recall  \\ 
\midrule
DeepLab-v3+ \cite{deeplabv3} & ResNet-50 & 59.3M & 47.35 & 48.44 \\ 
DeepLab-v3+ \cite{deeplabv3} & DRN-D-105 & 40.7M & 43.78 &  51.58 \\ 
DeepLab-v3+ \cite{deeplabv3} & Xception & 54.7M & 45.51 & 68.96 \\ 
DeepLab-v3+ \cite{deeplabv3} & MobileNet-V2 & 5.8M & 49.48 & 74.62 \\
AdapNet++ \cite{valada19ijcv} & EfficientNet & 16.3M & \textbf{56.57} & 77.73 \\
ENet \cite{paszke2016enet} & Custom & 0.4M & 45.90  & \textbf{83.14} \\ 
BiSeNet \cite{yu2018bisenet} & Xception & 5.8M & 43.79 & 73.85 \\ 
Fast-SCNN \cite{poudel2019fast} & Custom & 1.1M & 47.42 & 68.70 \\ 
\bottomrule
\end{tabular} 
\end{table}

In this section, we investigate the performance of various recently proposed semantic segmentation architectures for self-supervised learning in our framework. Larger networks such as DeepLabv3+~\cite{deeplabv3} typically have several millions of parameters which makes it not only inefficient for online robotics application but also require an enormous amount of computational capacity for training. Networks such as ERFNet~\cite{romera2017erfnet}, BiSeNet~\cite{yu2018bisenet}, and F-SCNN~\cite{poudel2019fast} rather focus on keeping the number of parameters low resulting in much higher training and inference speeds on a given hardware budget. For our self-supervised terrain segmentation framework, we evaluate four different backbones for the DeepLabv3+ architecture on our weakly supervised terrain classification dataset: Residual Network (ResNet-50)~\cite{he2016deep}, Dilated Residual Network (DRN-D-105)~\cite{yu2017dilated}, Depthwise Separable Convolutions~(Xception) \cite{chollet2017xception}, and MobileNetV2~\cite{sandler2018mobilenetv2}. Additionally, we evaluate three recent segmentation networks with a smaller number of parameters: ENet~\cite{paszke2016enet}, BiSeNet~\cite{yu2018bisenet}, and Fast-SCNN~\cite{poudel2019fast}. The results from this experiments are presented in \tabref{tab:ablation}. We observe that the AdapNet++ network yields the highest mean IoU score on our weakly labeled Freiburg Terrains dataset, while ENet achieves the highest recall. Therefore, we adopt the AdapNet++ architecture with the Efficient backbone for self-supervised semantic terrain segmentation in our framework.

\subsubsection{Influence of the Perspective Transformation}
\label{sec:perspective}

\begin{table}
\centering
\caption{Comparison of the self-supervised semantic terrain segmentation performance while learning from different perspectives.}
\label{tab:ablationperspective}
\begin{tabular}{p{3.2cm}p{1.2cm}}
\toprule
Perspective & mIoU ($\%$)  \\ 
\midrule
Standard & 41.89 \\ 
Birds-eye-view & \textbf{56.53} \\
\bottomrule
\end{tabular} 
\end{table}

\begin{figure*}
\centering
\footnotesize
\setlength{\tabcolsep}{0.1cm}
{\renewcommand{\arraystretch}{0.4}
    \begin{tabular}{P{0.5cm}P{4.1cm}P{4.1cm}P{4.1cm}P{4.1cm}}
        & Input Image & Output without Fine-Tuning & Output with Fine-Tuning \\
        \X{(a)} & \includegraphics[width=\linewidth,trim={2cm 2.5cm 2cm 2.5cm},clip]{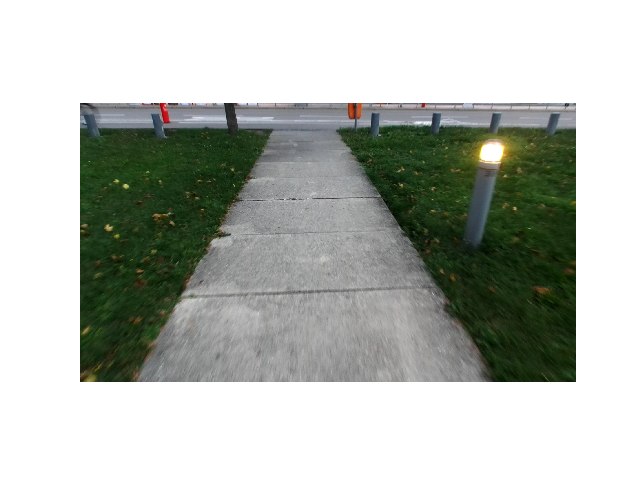} & \includegraphics[width=\linewidth,trim={2cm 2.5cm 2cm 2.5cm},clip]{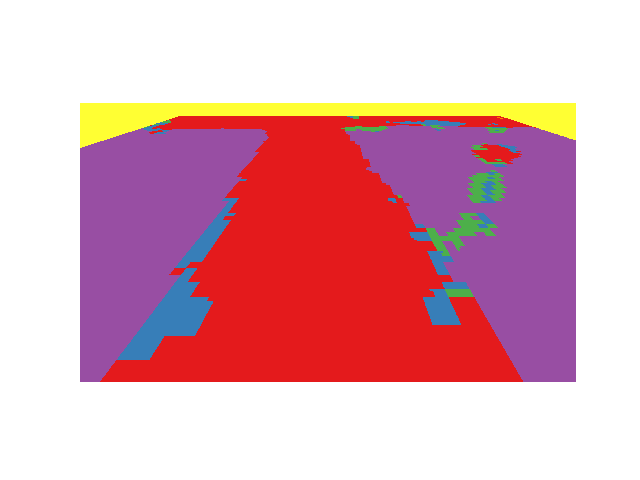} & \includegraphics[width=\linewidth,trim={2cm 2.5cm 2cm 2.5cm},clip]{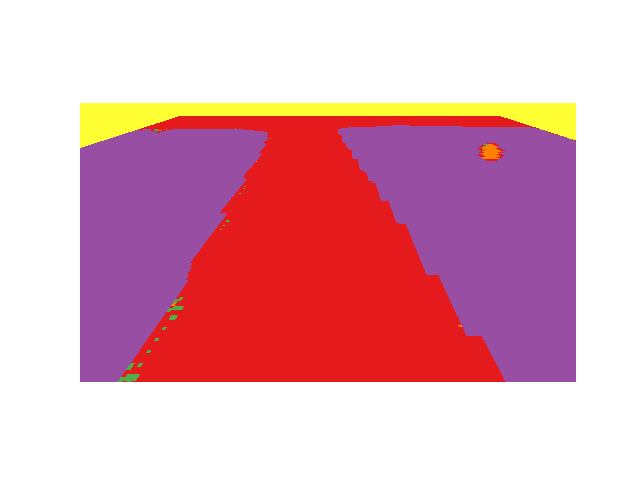} \\
        \\
        \X{(b)} & \includegraphics[width=\linewidth,trim={2cm 2.5cm 2cm 2.5cm},clip]{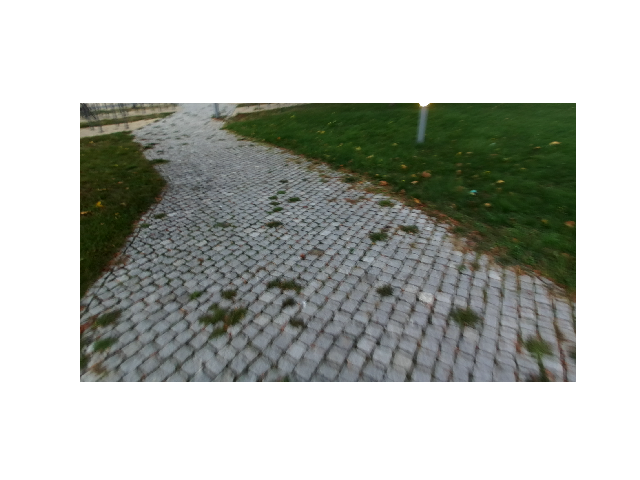} & \includegraphics[width=\linewidth,trim={2cm 2.5cm 2cm 2.5cm},clip]{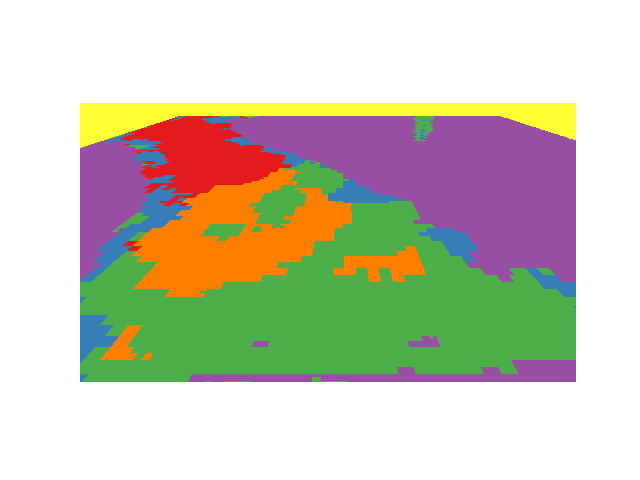} & \includegraphics[width=\linewidth,trim={2cm 2.5cm 2cm 2.5cm},clip]{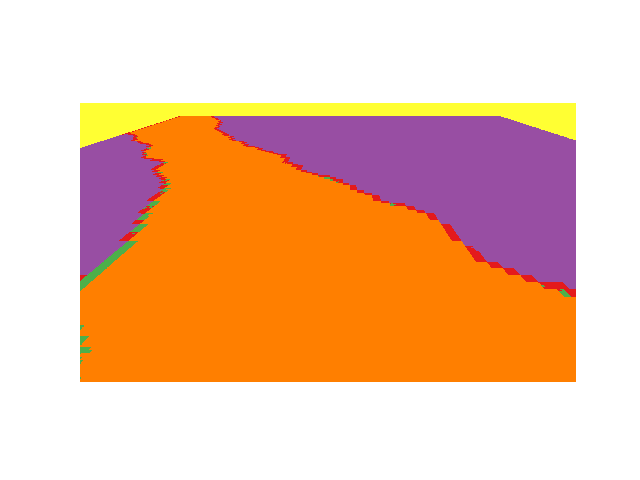}
    \end{tabular}}
    \caption{Qualitative results on a new low light dataset that was captured at dusk that has considerable amount of motion blur, color noise, and artificial lighting. We show a comparison between the terrain classification model without and with fine-tuning on training data created using our self-supervised approach. The legend for terrain labels correspond to the colors shown in \tabref{tab:iou-comparison}.}
    \label{fig:compare-models} 
\end{figure*}

Semantic segmentation is typically performed using a scene representation from the perspective of the camera. We investigate in the influence of a perspective transformation of the terrain images into a perspective according to the position of a virtual birds-eye-view camera. A birds-eye-view representation of a scene has the advantage that the terrain patterns that are visible in the image do not have any dependency on the distance of a patch of terrain to the camera. \tabref{tab:ablationperspective} shows the results from this experiment where we observe that the birds-eye-view terrain image representation yields a higher mean IoU score than from the standard perspective of the camera. This can be attributed to the fact that the typically occurring terrain-specific patterns are independent from the distance to the camera in the birds-eye-view perspective.

\subsection{Generalization to Terrain Appearance Changes}
\label{sec:generalization}

One of the major advantages of our self-supervised approach is that new labels on previously unseen terrains can easily be generated by the robot automatically. While the terrain traversal sounds do not substantially vary with the weather conditions other than rain and winds, the visual appearance of terrain can vary depending on several factors including time of day, season or cloudiness. In order to demonstrate the advantages of audio-based self-supervision, we collected an additional dataset in lighting conditions that are not present in the Freiburg Terrains dataset. We record data at dusk with low light conditions and artificial lighting resulting in a variation in terrain hues and substantial motion blur. We qualitatively compare the terrain classification results for a model trained exclusively on the Freiburg Terrains dataset, and a model trained jointly on the Freiburg Terrains dataset as well as on the new low light dataset. Qualitative results from this experiment is shown in \figref{fig:compare-models}.

Experiments demonstrate that our SE-R model trained on this combined dataset is able to automatically label $93.1\%$ of the audio clips in the new low-light dataset accurately. Our proposed framework enables us to fine-tune the semantic terrain segmentation model in a new domain without any human labeling effort, as the labels in the new domain can automatically be generated using the cluster indices of the audio sample embeddings. We observe that after fine-tuning the segmentation model for 10 additional epochs, it accurately predicts the terrains visible in the majority of the scenes. Whereas, the original model that was not fine-tuned does not generalize effectively to the visual appearances of terrains in the new scenes. The changes in lighting and hue of terrains in addition to motion blur and other artifacts of low light cannot be easily modeled as part of the augmentations that are performed while training. We argue that the ability to adapt to new environments in a self-supervised manner is an essential ability for autonomous robots to successfully navigate in changing environments and it brings us a step closer towards lifelong learning of traversability estimation.

\subsection{Semantic Terrain Mapping and Trajectory Planning}
\label{sec:planning}

\begin{figure*} 
\captionsetup[subfigure]{labelformat=empty}
    \centering
  \subfloat[(a) Birds-eye-view map of experimental area 1 \label{createdrgb}]{%
       \includegraphics[width=0.46\linewidth, 
       trim={0.5cm 0.5cm 8.5cm 5cm}, clip] {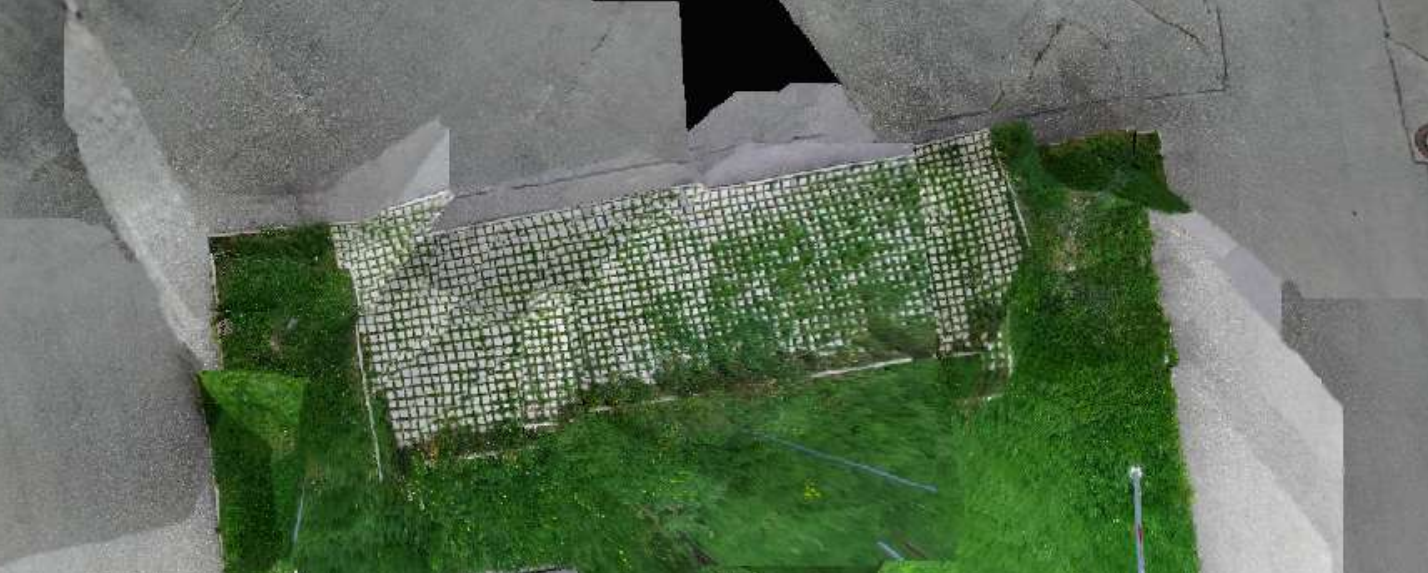}}
\hspace{1cm}
   \subfloat[(b) Semantic terrain map of experimental area 1 \label{bla}]{%
       \includegraphics[width=0.46\linewidth, 
       trim={0.5cm 0.5cm 8.5cm 5cm}, clip] {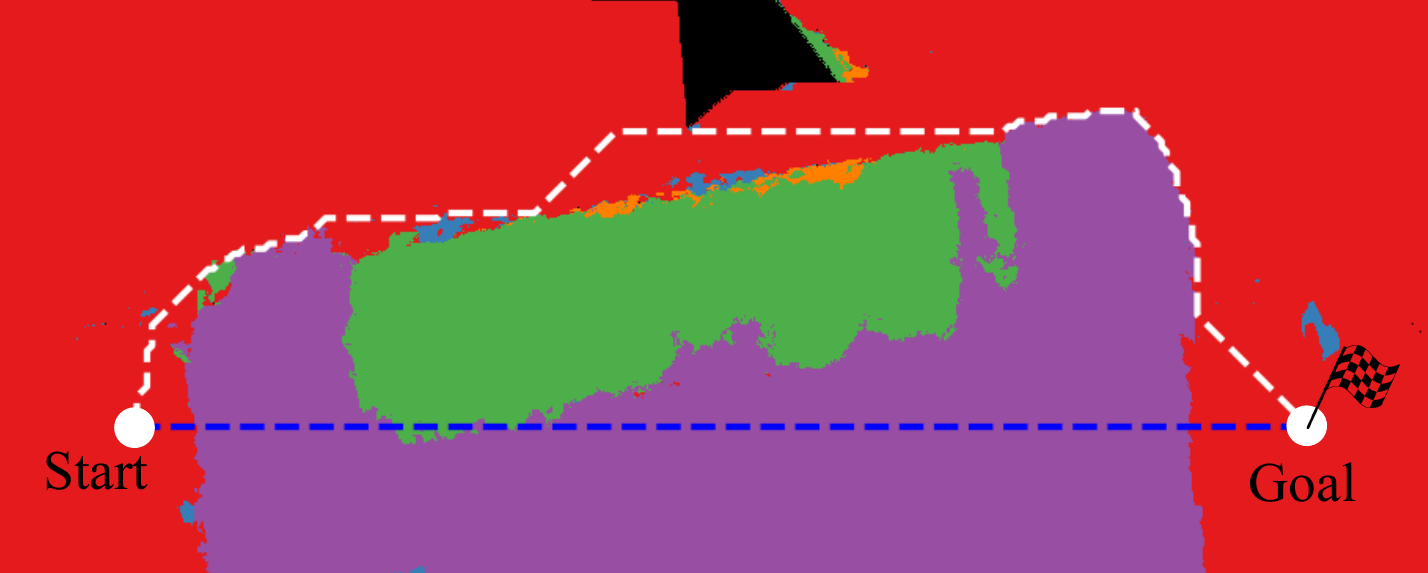}}\\
         \subfloat[(c) Birds-eye-view map of experimental area 2  \label{createdrgb2}]{%
       \includegraphics[width=0.46\linewidth, trim={1.5cm 3cm 8cm 1cm}, clip]{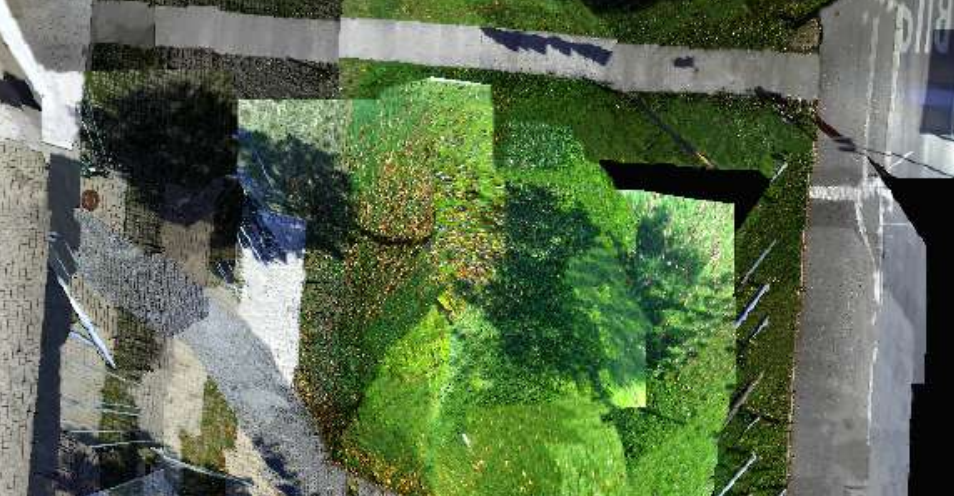}}
\hspace{1cm}
   \subfloat[(d) Semantic terrain map of experimental area 2 \label{bla2}]{%
       \includegraphics[width=0.46\linewidth, trim={1.5cm 3cm 8cm 1cm}, clip]{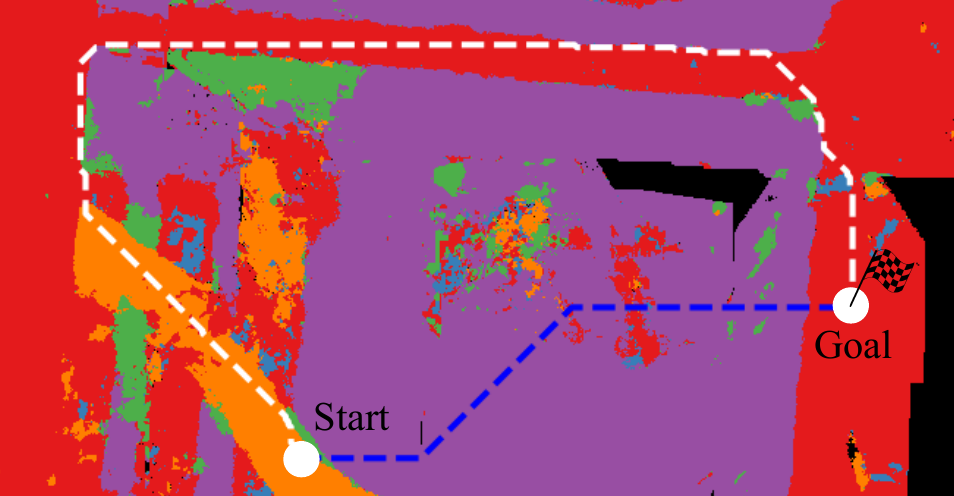}}
  \caption{Tiled birds-eye-view images (a, c) and the corresponding semantic terrain maps (b, d) built from the predictions of our self-supervised semantic terrain segmentation model. A cost-aware trajectory is shown as a white line and a cost-unaware trajectory is shown as a blue line. The legend for terrain labels correspond to the colors shown in \tabref{tab:iou-comparison}. Note how the cost-aware trajectory successfully avoids terrain classes \textit{Grass} and \textit{Parking Lot} that have higher traversal cost than \textit{Asphalt}.}
  \label{fig:map} 
\end{figure*}

In this section, we demonstrate the utility of our proposed self-supervised semantic segmentation framework for building semantic terrain maps of the environment. We further show how the generated terrain map enables a trajectory planner to choose lower cost trajectories compared to not using terrain-aware planning. As a proof of concept, we use the Dijkstra-algorithm for planning minimum cost trajectories. To build such a map, we use the poses of the robot that we obtain using our SLAM system and the terrain predictions of the birds-eye-view camera images. We project each image into the correct location in a global map using the 3-D camera pose and we use no additional image blending or image alignment optimization. For each single birds-eye-view image, we generate pixel-wise terrain classification predictions using our self-supervised semantic segmentation model. We then project these segmentation mask predictions into their corresponding locations in the global semantic terrain map, similar to the procedure that we employ for the birds-eye-view images. When there are predictions of a terrain location from multiple views, we choose the class with the highest prediction count for each pixel in the map. We also experimented with fusing the predictions from multiple views using Bayesian fusion which yields similar results.

\figref{fig:map} shows the tiled birds-eye-view of two small sections of an experimental site and the corresponding tiled semantic prediction map obtained using our model that also shows two exemplary trajectories. It can be observed that our self-supervised terrain segmentation model yields predictions that are for the most part globally consistent. In \figref{fig:map}~(b), we see that the class boundaries for the terrain classes such as \textit{Asphalt}, \textit{Grass}, and \textit{Parking Lot} are well-aligned with the ground truth terrains. It can also be seen that even the overgrown grass in the parking lot that blends into different terrains is being segmented. While in \figref{fig:map}~(d) we observe some misclassifications in the cobblestone area on the left side of the map and in the center of the grass region due to shadows from trees. However the terrain classes \textit{Grass} and \textit{Asphalt} are accurately predicted in the majority of pixels.

For trajectory planning, we interpret the semantic terrain map as a graph with each neighboring pixel pair as an edge and each pixel as a node. Edge costs are assigned according to their connected node class. We first plan a trajectory with equal cost assigned to each pixel which results in the trajectory shown in red. Note that the trajectory planned without any terrain awareness is the shortest route between points \textit{Start} and \textit{Goal} as this minimizes the global cost. In order to illustrate the advantages of cost-aware planning, we now associate high cost with the traversal of class \textit{Grass}, medium cost with \textit{Parking Lot}, and low cost with terrain \textit{Asphalt}. The trajectory planned with the aforementioned costs as constraints is shown in white. In \figref{fig:map}~(b), we can see that the trajectory avoids \textit{Grass} or the \textit{Parking Lot} altogether at the cost of a longer trajectory but more safely and efficiently traversable by our robot. While in \figref{fig:map}~(d), we can see that the planned terrain-aware trajectory again avoids high-cost terrains such as \textit{Grass} or \textit{Parking Lot} and follows \textit{Cobblestone} or \textit{Asphalt} to reach the goal. These experiments demonstrate that our framework can readily be used for planning efficient terrain-aware trajectories for robot navigation.
\section{Conclusion}
\label{sec:conclusions}

In this work, we proposed a self-supervised terrain classification framework that exploits the training signal from an unsupervised proprioceptive terrain classifier to learn an exteroceptive classifier for pixel-wise semantic terrain segmentation. We presented a novel heuristic for triplet sampling in metric learning that leverages a complementary modality as opposed to the typical strategy that requires ground truth labels. We employed this proposed heuristic for unsupervised clustering of vehicle-terrain interaction sound embeddings and subsequently used the resulting clusters formed by the audio embeddings for self-supervised labeling of terrain patches in images. We then trained a semantic terrain segmentation network from these weak labels for dense pixel-wise classification of terrains that are in front of the robot. 

We introduced the new challenging Freiburg Terrains dataset that we publicly released to encourage future work on self-supervised multimodal learning. We demonstrated the performance of our framework on this first-of-a-kind dataset where we showed that our proposed unsupervised proprioceptive classification approach outperforms existing unsupervised methods achieving state-of-the-art performance. We also presented our self-supervised semantic terrain segmentation results where our model achieves a comparable performance to training on fully supervised manually annotated labels. Additionally, we presented exhaustive qualitative results and ablation studies to highlight the importance of the contributions that we made in this work. Finally, we demonstrated the utility of our framework for semantic terrain mapping and trajectory planning for robot navigation. More importantly, we presented experiments in a new environment that show the generalization ability of our unsupervised proprioceptive terrain classifier and the ability of our semantic terrain segmentation model to automatically adapt to new environments in a self-supervised manner. We believe that this work has now brought us a step closer to being able to learn in a lifelong manner.

\bibliographystyle{IEEEtran}
\footnotesize
\bibliography{sections/references}

\begin{thebibliography}{10}
\providecommand{\url}[1]{#1}
\csname url@samestyle\endcsname
\providecommand{\newblock}{\relax}
\providecommand{\bibinfo}[2]{#2}
\providecommand{\BIBentrySTDinterwordspacing}{\spaceskip=0pt\relax}
\providecommand{\BIBentryALTinterwordstretchfactor}{4}
\providecommand{\BIBentryALTinterwordspacing}{\spaceskip=\fontdimen2\font plus
\BIBentryALTinterwordstretchfactor\fontdimen3\font minus
  \fontdimen4\font\relax}
\providecommand{\BIBforeignlanguage}[2]{{%
\expandafter\ifx\csname l@#1\endcsname\relax
\typeout{** WARNING: IEEEtran.bst: No hyphenation pattern has been}%
\typeout{** loaded for the language `#1'. Using the pattern for}%
\typeout{** the default language instead.}%
\else
\language=\csname l@#1\endcsname
\fi
#2}}
\providecommand{\BIBdecl}{\relax}
\BIBdecl

\bibitem{sofman2006improving}
B.~Sofman, E.~Lin, J.~A. Bagnell, J.~Cole, N.~Vandapel, and A.~Stentz,
  ``Improving robot navigation through self-supervised online learning,''
  \emph{Journal of Field Robotics}, vol.~23, no. 11-12, pp. 1059--1075, 2006.

\bibitem{hadsell2008deep}
R.~Hadsell, A.~Erkan, P.~Sermanet, M.~Scoffier, U.~Muller, and Y.~LeCun, ``Deep
  belief net learning in a long-range vision system for autonomous off-road
  driving,'' in \emph{IEEE/RSJ International Conference on Intelligent Robots
  and Systems}, 2008, pp. 628--633.

\bibitem{konolige2009mapping}
K.~Konolige, M.~Agrawal, M.~R. Blas, R.~C. Bolles, B.~Gerkey, J.~Sola, and
  A.~Sundaresan, ``Mapping, navigation, and learning for off-road traversal,''
  \emph{Journal of Field Robotics}, vol.~26, no.~1, pp. 88--113, 2009.

\bibitem{brooks2012self}
C.~A. Brooks and K.~Iagnemma, ``Self-supervised terrain classification for
  planetary surface exploration rovers,'' \emph{Journal of Field Robotics},
  vol.~29, no.~3, pp. 445--468, 2012.

\bibitem{ojeda2006terrain}
L.~Ojeda, J.~Borenstein, G.~Witus, and R.~Karlsen, ``Terrain characterization
  and classification with a mobile robot,'' \emph{Journal of Field Robotics},
  vol.~23, no.~2, pp. 103--122, 2006.

\bibitem{valada2018deep}
A.~Valada, L.~Spinello, and W.~Burgard, ``Deep feature learning for
  acoustics-based terrain classification,'' in \emph{Robotics Research}, 2018,
  pp. 21--37.

\bibitem{libby2012using}
J.~Libby and A.~J. Stentz, ``Using sound to classify vehicle-terrain
  interactions in outdoor environments,'' in \emph{IEEE International
  Conference on Robotics and Automation}, 2012, pp. 3559--3566.

\bibitem{valada2017deep}
A.~Valada and W.~Burgard, ``Deep spatiotemporal models for robust
  proprioceptive terrain classification,'' \emph{The International Journal of
  Robotics Research}, vol.~36, no. 13-14, pp. 1521--1539, 2017.

\bibitem{xie2019best}
C.~Xie, Y.~Xiang, A.~Mousavian, and D.~Fox, ``The best of both modes:
  Separately leveraging rgb and depth for unseen object instance
  segmentation,'' \emph{arXiv preprint arXiv:1907.13236}, 2019.

\bibitem{radwan2018multimodal}
N.~Radwan, A.~Valada, and W.~Burgard, ``Multimodal interaction-aware motion
  prediction for autonomous street crossing,'' \emph{arXiv preprint
  arXiv:1808.06887}, 2018.

\bibitem{blum2018modular}
H.~Blum, A.~Gawel, R.~Siegwart, and C.~Cadena, ``Modular sensor fusion for
  semantic segmentation,'' in \emph{2018 IEEE/RSJ International Conference on
  Intelligent Robots and Systems (IROS)}, 2018.

\bibitem{valada2016towards}
A.~Valada, G.~Oliveira, T.~Brox, and W.~Burgard, ``Towards robust semantic
  segmentation using deep fusion,'' in \emph{Robotics: Science and Systems (RSS
  2016) Workshop, Are the Sceptics Right? Limits and Potentials of Deep
  Learning in Robotics}, 2016.

\bibitem{happold2006enhancing}
M.~Happold, M.~Ollis, and N.~Johnson, ``Enhancing supervised terrain
  classification with predictive unsupervised learning,'' in \emph{Robotics:
  Science and Systems}, 2006.

\bibitem{hadsell2009learning}
R.~Hadsell, P.~Sermanet, J.~Ben, A.~Erkan, M.~Scoffier, K.~Kavukcuoglu,
  U.~Muller, and Y.~LeCun, ``Learning long-range vision for autonomous off-road
  driving,'' \emph{Journal of Field Robotics}, vol.~26, no.~2, 2009.

\bibitem{otsu2016autonomous}
K.~Otsu, M.~Ono, T.~J. Fuchs, I.~Baldwin, and T.~Kubota, ``Autonomous terrain
  classification with co-and self-training approach,'' \emph{IEEE Robotics and
  Automation Letters}, vol.~1, no.~2, pp. 814--819, 2016.

\bibitem{bekhti2014terrain}
M.~A. Bekhti, Y.~Kobayashi, and K.~Matsumura, ``Terrain traversability analysis
  using multi-sensor data correlation by a mobile robot,'' in \emph{IEEE/SICE
  International Symposium on System Integration}, 2014.

\bibitem{zhou2012self}
S.~Zhou, J.~Xi, M.~W. McDaniel, T.~Nishihata, P.~Salesses, and K.~Iagnemma,
  ``Self-supervised learning to visually detect terrain surfaces for autonomous
  robots operating in forested terrain,'' \emph{Journal of Field Robotics},
  vol.~29, no.~2, pp. 277--297, 2012.

\bibitem{stavens2012self}
D.~Stavens and S.~Thrun, ``A self-supervised terrain roughness estimator for
  off-road autonomous driving,'' \emph{arXiv preprint arXiv:1206.6872}, 2012.

\bibitem{wellhausen2019should}
L.~Wellhausen, A.~Dosovitskiy, R.~Ranftl, K.~T. Walas, C.~C. Lerma, and
  M.~Hutter, ``Where should i walk? predicting terrain properties from images
  via self-supervised learning,'' \emph{IEEE Robotics and Automation Letters},
  2019.

\bibitem{nava2019learning}
M.~Nava, J.~Guzzi, R.~O. Chavez-Garcia, L.~M. Gambardella, and A.~Giusti,
  ``Learning long-range perception using self-supervision from short-range
  sensors and odometry,'' \emph{IEEE Robotics and Automation Letters}, vol.~4,
  no.~2, pp. 1279--1286, 2019.

\bibitem{chopra2005learning}
S.~Chopra, R.~Hadsell, Y.~LeCun \emph{et~al.}, ``Learning a similarity metric
  discriminatively, with application to face verification,'' in \emph{CVPR},
  2005.

\bibitem{schroff2015facenet}
F.~Schroff, D.~Kalenichenko, and J.~Philbin, ``Facenet: A unified embedding for
  face recognition and clustering,'' in \emph{Proceedings of the IEEE
  conference on computer vision and pattern recognition}, 2015.

\bibitem{sohn2016improved}
K.~Sohn, ``Improved deep metric learning with multi-class n-pair loss
  objective,'' in \emph{Advances in Neural Information Processing Systems},
  2016.

\bibitem{xie2016unsupervised}
J.~Xie, R.~Girshick, and A.~Farhadi, ``Unsupervised deep embedding for
  clustering analysis,'' in \emph{International conference on machine
  learning}, 2016, pp. 478--487.

\bibitem{guo2017improved}
X.~Guo, L.~Gao, X.~Liu, and J.~Yin, ``Improved deep embedded clustering with
  local structure preservation.'' in \emph{IJCAI}, 2017, pp. 1753--1759.

\bibitem{barnes2017find}
D.~Barnes, W.~Maddern, and I.~Posner, ``Find your own way: Weakly-supervised
  segmentation of path proposals for urban autonomy,'' in \emph{IEEE
  International Conference on Robotics and Automation}, 2017.

\bibitem{hirose2018gonet}
N.~Hirose, A.~Sadeghian, M.~V{\'a}zquez, P.~Goebel, and S.~Savarese, ``Gonet: A
  semi-supervised deep learning approach for traversability estimation,'' in
  \emph{2018 IEEE/RSJ International Conference on Intelligent Robots and
  Systems (IROS)}, 2018, pp. 3044--3051.

\bibitem{kummerle2015autonomous}
R.~K{\"u}mmerle, M.~Ruhnke, B.~Steder, C.~Stachniss, and W.~Burgard,
  ``Autonomous robot navigation in highly populated pedestrian zones,''
  \emph{Journal of Field Robotics}, vol.~32, no.~4, pp. 565--589, 2015.

\bibitem{engel2017neural}
J.~Engel, C.~Resnick, A.~Roberts, S.~Dieleman, M.~Norouzi, D.~Eck, and
  K.~Simonyan, ``Neural audio synthesis of musical notes with wavenet
  autoencoders,'' in \emph{Proceedings of the 34th International Conference on
  Machine Learning-Volume 70}, 2017, pp. 1068--1077.

\bibitem{kuhn1955hungarian}
H.~W. Kuhn, ``The hungarian method for the assignment problem,'' \emph{Naval
  research logistics quarterly}, vol.~2, no. 1-2, pp. 83--97, 1955.

\bibitem{guo2017deep}
X.~Guo, X.~Liu, E.~Zhu, and J.~Yin, ``Deep clustering with convolutional
  autoencoders,'' in \emph{International Conference on Neural Information
  Processing}, 2017, pp. 373--382.

\bibitem{sandler2018mobilenetv2}
M.~Sandler, A.~Howard, M.~Zhu, A.~Zhmoginov, and L.-C. Chen, ``Mobilenetv2:
  Inverted residuals and linear bottlenecks,'' in \emph{Proceedings of the IEEE
  Conference on Computer Vision and Pattern Recognition}, 2018, pp. 4510--4520.

\bibitem{valada19ijcv}
A.~Valada, R.~Mohan, and W.~Burgard, ``Self-supervised model adaptation for
  multimodal semantic segmentation,'' \emph{International Journal of Computer
  Vision (IJCV)}, 2019.

\bibitem{deeplabv3}
L.~Chen, Y.~Zhu, G.~Papandreou, F.~Schroff, and H.~Adam, ``Encoder-decoder with
  atrous separable convolution for semantic image segmentation,'' \emph{arXiv
  preprint arXiv: 1802.02611}, 2018.

\bibitem{paszke2016enet}
A.~Paszke, A.~Chaurasia, S.~Kim, and E.~Culurciello, ``Enet: A deep neural
  network architecture for real-time semantic segmentation,'' \emph{arXiv
  preprint arXiv:1606.02147}, 2016.

\bibitem{yu2018bisenet}
C.~Yu, J.~Wang, C.~Peng, C.~Gao, G.~Yu, and N.~Sang, ``Bisenet: Bilateral
  segmentation network for real-time semantic segmentation,'' in
  \emph{Proceedings of the European Conference on Computer Vision}, 2018.

\bibitem{poudel2019fast}
R.~P. Poudel, S.~Liwicki, and R.~Cipolla, ``Fast-scnn: fast semantic
  segmentation network,'' \emph{arXiv preprint arXiv:1902.04502}, 2019.

\bibitem{romera2017erfnet}
E.~Romera, J.~M. Alvarez, L.~M. Bergasa, and R.~Arroyo, ``Erfnet: Efficient
  residual factorized convnet for real-time semantic segmentation,'' \emph{IEEE
  Transactions on Intelligent Transportation Systems}, 2017.

\bibitem{he2016deep}
K.~He, X.~Zhang, S.~Ren, and J.~Sun, ``Deep residual learning for image
  recognition,'' in \emph{Proceedings of the IEEE conference on computer vision
  and pattern recognition}, 2016, pp. 770--778.

\bibitem{yu2017dilated}
F.~Yu, V.~Koltun, and T.~Funkhouser, ``Dilated residual networks,'' in
  \emph{Proceedings of the IEEE conference on computer vision and pattern
  recognition}, 2017, pp. 472--480.

\bibitem{chollet2017xception}
F.~Chollet, ``Xception: Deep learning with depthwise separable convolutions,''
  in \emph{Proceedings of the IEEE conference on computer vision and pattern
  recognition}, 2017, pp. 1251--1258.

\end{thebibliography}

%

\begin{IEEEbiography}[{\includegraphics[width=1in,height=1.25in,clip,keepaspectratio]{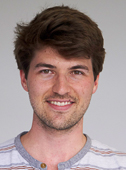}}]{Jannik Z\"urn}
  is a Ph.D.~student in the Autonomous Intelligent Systems group headed by Wolfram Burgard. He received his M.S.~degree from the Karlsruhe Institute of Technology in 2018. His research focuses on Deep Learning for robot perception and planning.
\end{IEEEbiography}

\begin{IEEEbiography}[{\includegraphics[width=1in,height=1.25in,clip,keepaspectratio]{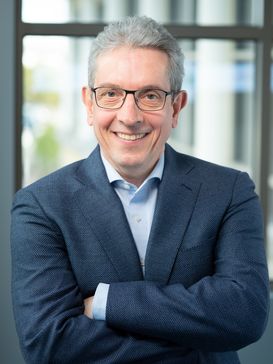}}]{Wolfram
    Burgard} is Vice President for Automated Driving Technology at the Toyota Research Institute in Los Altos, USA. He is on leave from a Professorship for Computer Science at the University of
  Freiburg, Germany where he heads the Laboratory for Autonomous
  Intelligent Systems. He received his Ph.D.~degree in computer
  science from the University of Bonn in 1991.  His areas of interest
  lie in robotics and artificial intelligence. In the past, Wolfram
  Burgard and his group developed several innovative probabilistic
  techniques for robot navigation and control. They cover different
  aspects including localization, mapping, path planning, and
  exploration. For his work, Wolfram Burgard received several best
  paper awards from outstanding national and international
  conferences. In 2009, Wolfram Burgard received the Gottfried Wilhelm
  Leibniz Prize, the most prestigious German research award. In 2010
  he received the Advanced Grant of the European Research
  Council. Wolfram Burgard is the spokesperson of the Cluster of
  Excellence BrainLinks-BrainTools, President of the IEEE Robotics and Automation Society, and fellow of the AAAI, EurAi and IEEE.\looseness=-1
\end{IEEEbiography}

\begin{IEEEbiography}[{\includegraphics[width=1in,height=1.25in,clip,keepaspectratio]{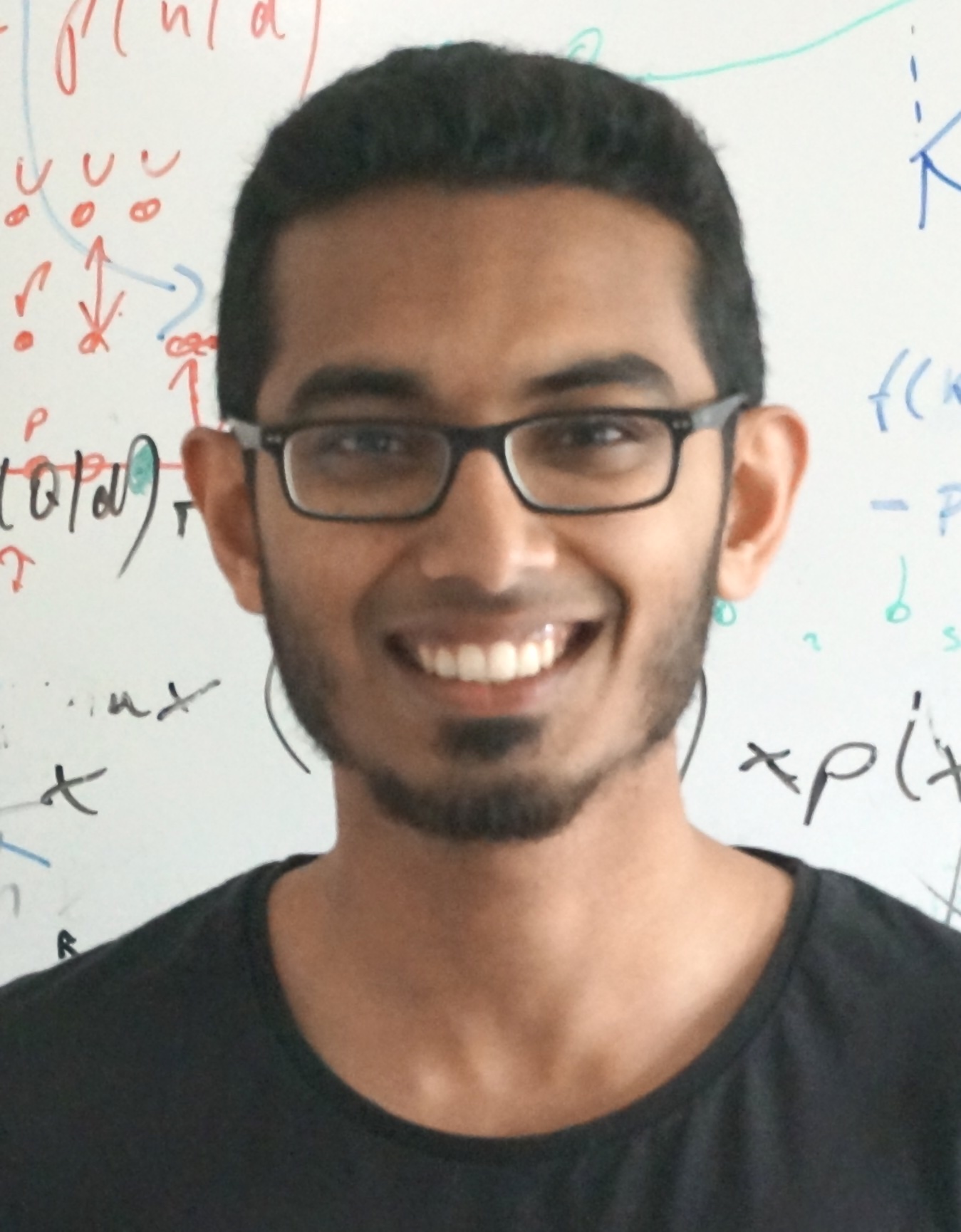}}]{Abhinav Valada}
is an Assistant Professor at the University of Freiburg, Germany where he heads the Robot Learning Laboratory. He received his Ph.D.~degree in Computer Science from the University of Freiburg in 2019 and his M.S.~degree in Robotics from Carnegie Mellon University in 2013. His research interests include robot learning, perception, state estimation, navigation, self-supervised and unsupervised learning.
\end{IEEEbiography}



\vfill


\end{document}